\pdfoutput=1
\documentclass[11pt]{article}

\usepackage[]{acl}

\usepackage{times}
\usepackage{latexsym}

\usepackage[T1]{fontenc}
\usepackage[utf8]{inputenc}

\usepackage{microtype}

\usepackage{inconsolata}

\usepackage{booktabs}
\usepackage{longtable}
\usepackage{tcolorbox}
\usepackage{stackengine}
\usepackage{multirow}
\usepackage{graphicx}
\usepackage{wrapfig}
\usepackage{hyperref}       % hyperlinks
\usepackage{url}            % simple URL typesetting
\usepackage{amsfonts}       % blackboard math symbols
\usepackage{nicefrac}       % compact symbols for 1/2, etc.
\usepackage{microtype}      % microtypography
\usepackage{xcolor,colortbl}         % colors
\usepackage{times}
\usepackage{latexsym}
\usepackage{multicol}
\usepackage{blindtext}
\usepackage{tabu}
\usepackage{amsmath, bm}

\usepackage{subcaption}
\usepackage{caption}
\usepackage[normalem]{ulem}
\usepackage{soul}
\usepackage[shortlabels]{enumitem}
\usepackage{array}
\usepackage{pgffor}
\usepackage{textcomp}
\usepackage{amssymb}
\usepackage{pifont}
\newcommand{\xmark}{\ding{55}}
\definecolor{lightgreen}{rgb}{0.8, 0.95, 0.8}
\definecolor{lightred}{rgb}{0.95, 0.8, 0.8}
\definecolor{naplesyellow}{rgb}{0.98, 0.85, 0.37}
\definecolor{pastelyellow}{rgb}{0.99, 0.99, 0.59}

\title{Multi-LogiEval: Towards Evaluating Multi-Step Logical Reasoning Ability of Large Language Models}

\author{Nisarg Patel$^*$ \quad Mohith Kulkarni$^*$ \quad Mihir Parmar$^*$ \quad Aashna Budhiraja \\ \textbf{Mutsumi Nakamura} \quad \textbf{Neeraj Varshney} \quad \textbf{Chitta Baral} \\\\ Arizona State University, USA\\
\small{\texttt{\{nppatel7, mkulka20, mparmar3, chitta\}@asu.edu}}}

\begin{document}
\maketitle

\begin{abstract}

As Large Language Models (LLMs) continue to exhibit remarkable performance in natural language understanding tasks, there is a crucial need to measure their ability for human-like multi-step logical reasoning. Existing logical reasoning evaluation benchmarks often focus primarily on simplistic single-step or multi-step reasoning with a limited set of inference rules. Furthermore, the lack of datasets for evaluating non-monotonic reasoning represents a crucial gap since it aligns more closely with human-like reasoning. To address these limitations, we propose \textit{Multi-LogiEval}, a comprehensive evaluation dataset encompassing multi-step logical reasoning with various inference rules and depths. \textit{Multi-LogiEval} covers three logic types — propositional, first-order, and non-monotonic consisting of more than 30 inference rules and more than 60 of their combinations with various depths. Leveraging this dataset, we conduct evaluations on a range of LLMs including GPT-4, ChatGPT, Gemini-Pro, Yi, Orca, and Mistral, employing a zero-shot chain-of-thought. Experimental results show that there is a significant drop in the performance of LLMs as the reasoning steps/depth increases (average accuracy of $\sim68\%$ at depth-1 to $\sim43\%$ at depth-5). We further conduct a thorough investigation of reasoning chains generated by LLMs which reveals several important findings. We believe that \textit{Multi-LogiEval} facilitates future research for evaluating and enhancing the logical reasoning ability of LLMs\footnote{Data is available at 
\url{https://github.com/Mihir3009/Multi-LogiEval}}.

%\url{https://anonymous.4open.science/r/Multi-LogiEval-FFDB}}.
% \maketitle
\def\thefootnote{*}\footnotetext{Equal Contribution}\def\thefootnote{\english{footnote}}

% Experimental results reveal several interesting findings such as \mihir{Add 1-2 solid findings from results.}
%Multi-LogiEval combines classical and non-classical logic, presenting a framework to emulate human-like multi-step reasoning across a diverse spectrum of 25 inference rules. 

%https://github.com/Mihir3009/Multi-LogiEval

\end{abstract}

\section{Introduction}
\label{sec:intro}

\begin{figure}
    \centering
    \includegraphics[width=\linewidth]{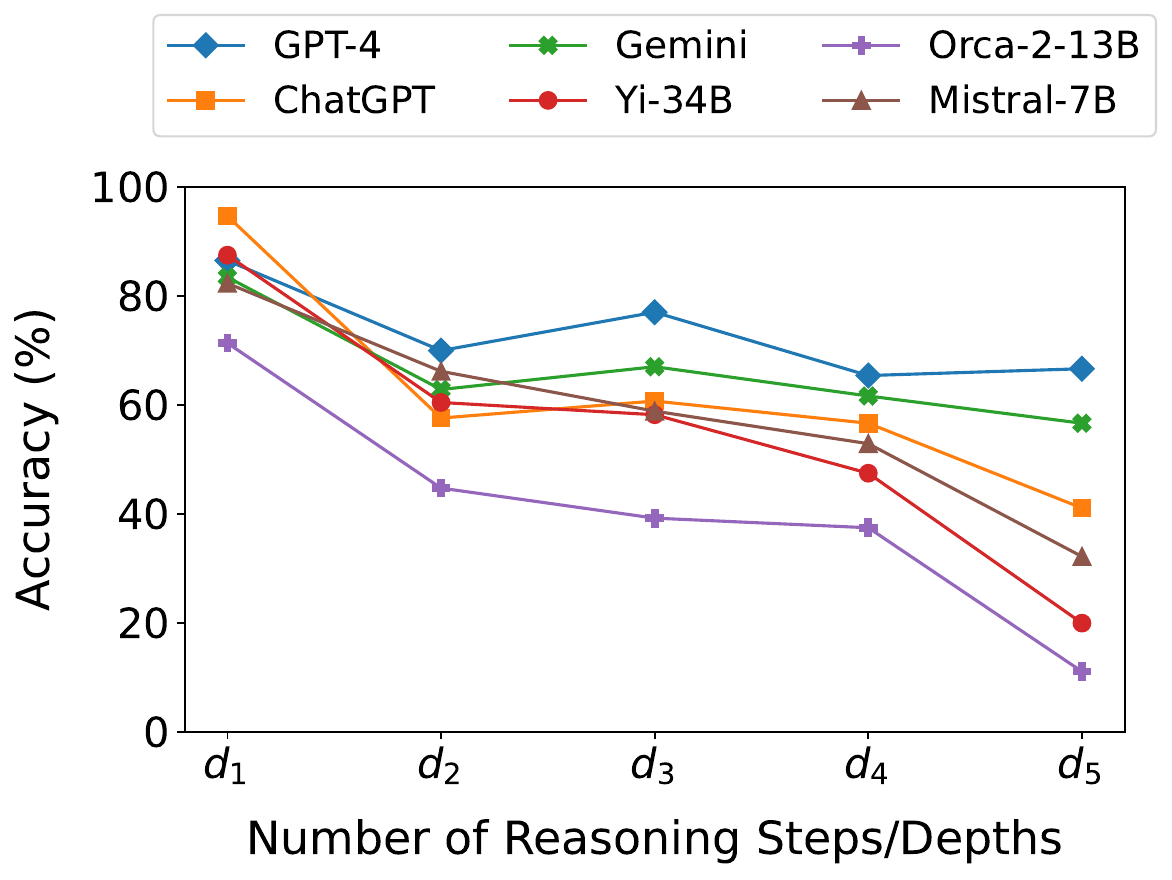}
    \caption{Performance (avg. accuracy across each depth for PL \& FOL) of various LLMs on \textit{Multi-LogiEval}.}
    \label{fig:teaser_figure}
\end{figure}

The ability to perform multi-step reasoning -- drawing conclusions from provided multiple premises -- is a hallmark of human intelligence.  
Recently, Large Language Models (LLMs) such as GPT-4, ChatGPT, Gemini, and Mistral \cite{jiang2023mistral} have achieved impressive performance on a variety of language tasks that were previously thought to be exclusive to humans \cite{openai2023gpt4, NEURIPS2020_1457c0d6, zhao2023survey}. However, the ability of these LLMs to perform multi-step logical reasoning over natural language remains under-explored, despite its various real-world applications \cite{khashabi2019reasoning, beygi-etal-2022-logical}. Although several datasets have been proposed \cite{luo2023towards} to evaluate the logical reasoning capabilities of LLMs, these datasets are limited in their scope by (1) evaluating simplistic single-step logical reasoning such as ProntoQA \cite{saparov2023language} and (2) evaluating multi-step logical reasoning, but only on a single type of logic and covering only a few logical inference rules as done in FOLIO \cite{han2022folio} and ProofWriter \citep{tafjord-etal-2021-proofwriter}. Furthermore, there are only a few benchmarks, such as LogicBench \cite{parmar2024logicbench} and BoardgameQA \cite{kazemi2023boardgameqa}, that cover reasoning such as non-monotonic which is closer to human-like reasoning. Motivated by this, our work aims to bridge these gaps by creating a more comprehensive and logically complex evaluation dataset by incorporating varying numbers of reasoning depths (i.e., multi-steps) to reach conclusions. In addition, past attempts have been made to evaluate multi-hop reasoning of language models \cite{mavi2022survey}. In contrast, our work systematically evaluates multi-hop logical reasoning over various inference rules and their combinations.

%mirroring the multi-step deductive reasoning that happens in the real world
%logical reasoning being necessary for
%where we achieve logical complexity

To this end, we propose \textit{Multi-LogiEval}, a systematically created Question-Answering (QA) dataset covering multi-step logical reasoning across three different logic types: Propositional Logic (PL), First-Order Logic (FOL), and Non-Monotonic (NM) reasoning. Our objective is to present a preliminary analysis of the LLMs’ ability to perform multi-step logical reasoning and demonstrate their failures even when performing simple reasoning. We believe that, regardless of whether such reasoning is available in some existing natural data (e.g., examinations), LLMs should do proper logical reasoning. Thus, we systematically compiled data using various inference rules and varying numbers of reasoning depths. In particular, our proposed dataset provides $\sim1.6k$ high-quality instances that cover 33 inference rules and reasoning patterns and more than 60 complex combinations of these inference rules with a different number of reasoning steps ($1\sim5$). Our choice of inference rules is further explained in section \ref{sec 3:para:choice of inference rules}. To evaluate LLMs, we formulate a binary classification task in \textit{Multi-LogiEval} where the context represents a natural language story consisting of logical statements, and the models have to determine whether the story logically entails a conclusion given in the question. Examples of instances are presented in Table \ref{tab:depth_examples_NL}. To develop \textit{Multi-LogiEval}, we propose a two-stage procedure: (i) creating meaningful combinations of inference rules to generate data instances with different reasoning depths, and (ii) prompt LLMs to generate \textit{<context, question, answer>} triplets consisting of different `ontologies' (i.e., a collection of concepts such as car, person, and animals). In the end, we perform human validation of each generated instance to ensure the quality.

We evaluate a range of LLMs, including GPT-4, ChatGPT, Gemini-Pro, Yi-34B \cite{ai2024yi}, Orca-2-13B \cite{mitra2023orca}, and Mistral-7B \cite{wei2021finetuned} on \textit{Multi-LogiEval} using Zero-shot Chain-of-Thought (Zero-shot-CoT) prompting \cite{wei2022chain}.
The zero-shot CoT approach allows us to determine LLM's ability to do logical reasoning based on parametric knowledge (acquired during pre-training) since we can not expect in-context examples of inference rules for various reasoning depths will always be available in prompts.
We measure the accuracy of LLMs' predictions on the binary classification task.
As illustrated in Figure \ref{fig:teaser_figure}, our experimental results indicate that LLMs performance decreases as the depth of reasoning increases, indicating mistakes in the initial reasoning step propagate further in the reasoning chain. 
The rationale behind the choice of binary classification task is that it provides systematic standard metric-based evaluation (i.e., direct comparison of LLMs’ performance in terms of accuracy), which could be more challenging with open-ended question-answer formats.
However, we also provide a manual and thorough analysis of the reasoning chain generated by LLMs revealing several findings such as the importance of contextual information, the lack of correlation between longer reasoning chains and better outcomes, and the lower performance of larger-scale open-source LLMs compared to smaller ones.

%Thus, we believe that \textit{Multi-LogiEval} facilitates future research for evaluating the multi-step logical reasoning ability of existing and upcoming LLMs. 
% In summary, our contributions are:

% \begin{enumerate}[noitemsep,topsep=0pt,leftmargin=*]
%     \item We propose Multi-LogiEval, a comprehensive multi-step logical reasoning benchmark by creating meaningful combinations of inference rules to create data instances with different reasoning depths
%     \item We introduce a novel data generation method for creating data instances with multi-step logical reasoning, closer to real-world scenarios, and application of prompting techniques to ensure the validation and quality of the generated data.
%     \item \mihir{add one/two sentences about evaluation and findings}
% \end{enumerate}

\section{Related Work}

\begin{table}[!bpt]
  \centering
  \resizebox{\linewidth}{!}{
    \begin{tabular}{c|c|c|c|c}
    \toprule
    \multicolumn{1}{c|}{\multirow{2}[0]{*}{Dataset}} & \multicolumn{3}{c|}{Logic Covered} & \multicolumn{1}{c}{\multirow{2}[1]{*}{\begin{tabular}[c]{@{}c@{}} Multi-Step\\Logical Reasoning \end{tabular}}} \\ \cmidrule{2-4} 
    & PL & FOL & NM & \\ \midrule \midrule
    
    % GSM8k  &  \cellcolor{lightred} \xmark  &  \cellcolor{lightred} \xmark     &   \cellcolor{lightred} \xmark    &  $\sim$  & \cellcolor{lightred} \xmark & human annotated\\

    % Ruletaker & \cellcolor{lightred} \xmark    &   \cellcolor{lightgreen} \checkmark    &   \cellcolor{lightred}\xmark   & \cellcolor{lightred}\xmark  \\
    
    LogicNLI   &  \cellcolor{lightred}\xmark   &  \cellcolor{lightgreen} \checkmark     &   \cellcolor{lightred}\xmark &  \cellcolor{lightred}\xmark % (They focus on Statistics of commented part)  
    \\ % ∧ ∨ ¬ → ≡ ∀ ∃
    
    ProofWriter     &   \cellcolor{lightgreen} \checkmark    &   \cellcolor{lightgreen} \checkmark    &   \cellcolor{lightred}\xmark   & \cellcolor{lightgreen}\checkmark  \\
    
    FOLIO     &    \cellcolor{lightred}\xmark   &  \cellcolor{lightgreen} \checkmark     &   \cellcolor{lightred}\xmark   &  \cellcolor{lightgreen}\checkmark  \\
    
    SimpleLogic     &    \cellcolor{lightgreen}\checkmark   &   \cellcolor{lightred}\xmark    &   \cellcolor{lightred}\xmark   &  \cellcolor{lightgreen}\checkmark %(They used a rule for FOL for generation) 
    \\
    
    ProntoQA     &  \cellcolor{lightred}\xmark   &  \cellcolor{lightgreen} \checkmark     &   \cellcolor{lightred}\xmark   &  \cellcolor{lightred}\xmark \\

    RuleTaker     &  \cellcolor{lightred}\xmark   &  \cellcolor{lightgreen} \checkmark     &   \cellcolor{lightred}\xmark   &  \cellcolor{lightgreen} \checkmark \\
    
    LogicBench &  \cellcolor{lightgreen}\checkmark    & \cellcolor{lightgreen}\checkmark     & \cellcolor{lightgreen}\checkmark     & \cellcolor{lightred}\xmark \\ \midrule

    \textbf{\textit{Multi-LogiEval}} &  \cellcolor{lightgreen}\checkmark    & \cellcolor{lightgreen}\checkmark     & \cellcolor{lightgreen}\checkmark     & \cellcolor{lightgreen}\checkmark \\ \bottomrule
    \end{tabular}%
    }
    \caption{Comparison of \textit{Multi-LogiEval} with existing datasets and benchmarks}
  \label{tab:dataset_comparison}%
\end{table}

\begin{table*}[htbp]
  \centering
  \resizebox{\linewidth}{!}{
    \begin{tabular}{c|c|c} \toprule
    \textbf{Rule} & \textbf{Propositional Logic} & \textbf{First-order Logic} \\ \midrule \midrule
    \begin{tabular}[c]{@{}c@{}}MP\end{tabular}  & $((p \to q) \land p) \vdash q $ & \begin{tabular}[c]{@{}c@{}}$(\forall x(p(x) \to q(x)) \land p(a)) \vdash q(a) $ \end{tabular} \\ \midrule
    
    \begin{tabular}[c]{@{}c@{}}MT\end{tabular} &  $((p\to q)\land \neg q)\vdash \neg p$ &  \begin{tabular}[c]{@{}c@{}}$(\forall x(p(x) \to q(x)) \land \neg q(a)) \vdash \neg p(a) $ \end{tabular} \\ \midrule
    
    \begin{tabular}[c]{@{}c@{}}HS\end{tabular} & $((p\to q))\land (q\to r))\vdash (p\to r)$ & \begin{tabular}[c]{@{}c@{}}$(\forall x((p(x) \to q(x)) \land (q(x) \to r(x))) \vdash (p(a) \to r(a)) $ \end{tabular} \\ \midrule
    
    \begin{tabular}[c]{@{}c@{}}DS\end{tabular} & $((p\lor q)\land \neg p)\vdash q$ & \begin{tabular}[c]{@{}c@{}}$(\forall x(p(x) \lor q(x)) \land \neg p(a)) \vdash q(a) $\end{tabular} \\ \midrule
    
    \begin{tabular}[c]{@{}c@{}}CD\end{tabular} & $((p\to q)\land (r\to s)\land (p\lor r))\vdash (q\lor s)$ & \begin{tabular}[c]{@{}c@{}}$(\forall x((p(x) \to q(x)) \land  (r(x) \to s(x))) \land (p(a) \lor r(a)))\ \vdash (q(a) \lor s(a)) $\end{tabular} \\ \midrule
    
    \begin{tabular}[c]{@{}c@{}}DD\end{tabular} & $((p\to q)\land (r\to s)\land (\neg q\lor \neg s))\vdash (\neg p\lor \neg r)$ & \begin{tabular}[c]{@{}c@{}}$(\forall x((p(x) \to q(x)) \land  (r(x) \to s(x))) \land (\neg q(a) \lor \neg s(a)))\ \vdash (\neg p(a) \lor \neg r(a)) $\end{tabular} \\ \midrule
    
    \begin{tabular}[c]{@{}c@{}}BD\end{tabular} & $((p\to q)\land (r\to s)\land (p\lor \neg s))\vdash (q\lor \neg r)$ & \begin{tabular}[c]{@{}c@{}}$(\forall x((p(x) \to q(x)) \land  (r(x) \to s(x))) \land (p(a) \lor \neg s(a)))\ \vdash (q(a) \lor \neg r(a)) $\end{tabular} \\ 
    \midrule
    
    \begin{tabular}[c]{@{}c@{}}CT\end{tabular}  & $(p\lor q)\dashv\vdash (q\lor p)$ &  \begin{tabular}[c]{@{}c@{}}$\forall x(p(x) \lor q(x)) \dashv\vdash \forall x(q(x) \lor p(x))$\end{tabular} \\
    \midrule
    \begin{tabular}[c]{@{}c@{}}DMT\end{tabular}  & $\neg (p\land q)\dashv\vdash \neg p\lor \neg q$ &  \begin{tabular}[c]{@{}c@{}}$\neg\forall x (p(x) \land q(x)) \dashv\vdash \exists x(\neg p(x) \lor \neg q(x))$\end{tabular} \\
    \midrule
    \begin{tabular}[c]{@{}c@{}}CO\end{tabular}  & $((p\to q)\land (p\to r)\vdash (p\to (q\land r)$ &  \begin{tabular}[c]{@{}c@{}}$\forall x ((p(x)\to q(x)) \land (p(x)\to r(x)))\vdash \forall x (p(x) \to (q(x) \land r(x)))$\end{tabular} \\
    \midrule
    \begin{tabular}[c]{@{}c@{}}IM\end{tabular}  & $(p\to (q\to r))\dashv\vdash ((p\land q) \to r)$ &  \begin{tabular}[c]{@{}c@{}}$\forall x(p(x)\to (q(x)\to r(x)))\dashv\vdash \forall x((p(x)\land q(x)) \to r(x))$\end{tabular} \\
    \midrule
    \begin{tabular}[c]{@{}c@{}}MI\end{tabular}  & $(p\to q)\dashv\vdash (\neg p \lor q)$ &  \begin{tabular}[c]{@{}c@{}} - \end{tabular} \\
    \midrule
    \begin{tabular}[c]{@{}c@{}}EG\end{tabular}  & - &  \begin{tabular}[c]{@{}c@{}}$ p(a) \vdash \exists x(p(x))$\end{tabular} \\
    \midrule
    \begin{tabular}[c]{@{}c@{}}UI\end{tabular}  & - &  \begin{tabular}[c]{@{}c@{}}$ \forall x(p(x)) \vdash p(a)$\end{tabular} \\
    
    % \begin{tabular}[c]{@{}c@{}}MI\end{tabular} &  $(p\to q)\vdash (\neg p\lor q)$ & - \\ \midrule
    
    % \begin{tabular}[c]{@{}c@{}}EI\end{tabular} & - &  ${\displaystyle \exists xP\left({x}\right) \Rightarrow P\left({a}\right)}$ \\ \midrule
    
    % \begin{tabular}[c]{@{}c@{}}UI\end{tabular} & - & $\displaystyle \forall x\,A\Rightarrow A\{x\mapsto a\}$ 
    % \\ 
    \bottomrule
    
    \end{tabular}%
}
    \caption{Inference rules that establish the relationship between premises and their corresponding conclusions. A subset of these inference rules is adapted from \citet{parmar2024logicbench}. MP: Modus Ponens, MT: Modus Tollens, HS: Hypothetical Syllogism, DS: Disjunctive Syllogism, CD: Constructive Dilemma, DD: Destructive Dilemma, BD: Bidirectional Dilemma, CT: Commutation, DMT: De Morgan's Theorem, CO: Composition, IM: Importation, MI: Material Implication, EG: Existential Generalization, UI: Universal Instantiation}
    \label{tab:classical_inference_rules}%
\end{table*}
Past attempts have been made to assess the logical reasoning ability of language models. For instance, LogiQA \citep{liu2021logiqa} and ReClor \citep{yureclor} evaluate diverse forms of logical reasoning by compiling multi-choice questions from standardized examinations, including multi-step reasoning. However, in contrast to our \textit{Multi-LogiEval}, these datasets involve mixed forms of reasoning and do not focus on assessing logical reasoning independently. Past attempts have been made to create datasets focusing on logical reasoning\citep{luo2023towards}. In terms of task formulation, our proposed dataset is similar to ProofWriter \citep{tafjord-etal-2021-proofwriter}, RuleTaker \cite{clark2021transformers}, FOLIO \citep{han2022folio}, ProntoQA \citep{saparov2023language}, and LogicBench \cite{parmar2024logicbench} which are QA datasets designed to evaluate logical reasoning ability independently. ProofWriter provides multi-hop proofs for each example, RuleTaker mainly covers the simple implication rules such as modus ponens, while FOLIO gives diverse and complex logical expressions and covers multi-step reasoning. However, it is only limited to FOL. ProntoQA \citep{saparov2023language} provides a QA dataset with explanation and reasoning steps but is limited to single-step modus ponens in FOL. Although LogicBench \cite{parmar2024logicbench} covers various inference rules and reasoning patterns comprehensively, it only contains single-step logical reasoning (see Table \ref{tab:dataset_comparison} for comparison). Additional datasets for evaluating multi-step logical reasoning also exist, such as SimpleLogic \citep{zhang2022paradox}, which only covers modus ponens inference rule, and RuleBert \citep{saeed-etal-2021-rulebert} which covers only soft logical rules. In contrast, \textit{Multi-LogiEval} evaluates logical reasoning independently beyond modus ponens. In addition, FLD (Formal Logic Deduction) \cite{morishita2023learning} has formal logic theory-based inference rules, and their combinations to create multi-step reasoning, but limited to PL and FOL. However, \textit{Multi-LogiEval} offers a broader set of inference rules for PL and FOL, along with their meaningful combinations for multi-step reasoning, in addition to NM reasoning.

%beyond PL and FOL
%(Formal Logic Deduction)
%do not evaluate logical reasoning independently

% A few past attempts have been made to create datasets to evaluate only logical reasoning while excluding other forms of reasoning. For example, CLUTTER \citep{sinha-etal-2019-clutrr} covers inductive reasoning, \cite{hahn2021teaching} covers temporal logic, and Ruletaker \citep{clark2021transformers} evaluates whether a transformer-based model emulates deductive reasoning over synthetically generated statements in a limited setting. LogicNLI \citep{tian-etal-2021-diagnosing} introduced a diagnostic benchmark for FOL reasoning, with the dataset constructed by automatically generating logic expressions and replacing the entity and attribute placeholders.

%Nevertheless, several crucial attributes motivated us to create \textit{Multi-LogiEval} (see Table for comparison). 

\section{Multi-LogiEval}
% The goal of \textit{Multi-LogiEval} is to provide multi-step logical reasoning instances for evaluating both existing and upcoming LLMs. 

% The selection of inference rules and reasoning patterns for our dataset is motivated by LogicBench collection \cite{parmar2024logicbench}. 
In developing \textit{Multi-LogiEval}, we leverage the capabilities of LLMs while employing different methods to generate data for NM compared to PL and FOL since the formulations for PL and FOL differ from NM. In particular, our data creation process consists of two major stages: (i) Generation of rule combination and (ii) Generation of data instances. 

\paragraph{Generation of rule combination} We create a meaningful combination of inference rules to achieve reasoning depths and define the complex question for each combination that will require multiple reasoning steps to answer. Here, each step generally corresponds to one inference rule.

\paragraph{Generation of data instances} Using the combinations of inference rules generated in the above step, we prompt the LLM to generate a more human-like natural language story embedded with logical rules as a context and then the following complex reasoning question. In this way, we generate data in the form of \textit{<context, question>} pairs for each combination of inference rules at each depth. 

% We have created the data for three different branches of logic: Propositional logic, First-order logic, and Non-monotonic logic. We use a similar approach for propositional logic and first-order logic, which are considered classical logic, and a different approach for non-monotonic logic, which is considered non-classical.
% % A similar approach is taken for both propositional and first-order logic, which are considered classical logic. This is because both branches of logic contain the inference rules, the only difference being first-order logic includes the rule combination generation stage.
% In the following sections, we explain the data creation process for both these types of logic.
\def\stackalignment{l}
\newcommand{\stackFour}[4]{
  \stackunder[5pt]{#1}{\stackunder[5pt]{#2}{\stackunder[5pt]{#3}{#4}}}
}
\newcommand{\stackFive}[5]{
    \stackunder{#1}{\stackunder{#2}{\stackunder{#3}{\stackunder{#4}{#5}}}}
}
\begin{table*}[htbp]
\centering
\resizebox{0.84\linewidth}{!}{
\begin{tabular}{cllcc}
\toprule
\textbf{Depth} & \textbf{Rule Combinations} & \textbf{Premises in Story} & \textbf{Premise in Question} & \textbf{Answer} \\
\midrule \midrule
1 &
\textbf{MT: }(P $\to$ Q) $\land$ $\lnot$Q  $\vdash$ $\lnot$P
&
(P $\to$ Q) 
& $\lnot$Q & 
\textbf{$\lnot$P: \checkmark}
% Is R true? \textbf{: Yes}
\\ \midrule
2 &
\begin{tabular}[c]{@{}l@{}} \textbf{MT: }(P $\to$ Q) $\land$ $\lnot$Q  $\vdash$ $\lnot$P \\
\textbf{DS: } (P $\lor$ R) $\land$ $\lnot$P $\vdash$ R 
\end{tabular}
&
(P $\lor$ R), (P $\to$ Q) 
& $\lnot$Q & 
\textbf{R: \checkmark}
% Is R true? \textbf{: Yes}

\\ \midrule

3 &
\begin{tabular}[c]{@{}l@{}} \textbf{HS: }(P $\to$ Q) $\land$ (Q $\to$ R) $\vdash$ (P $\to$ R) \\ \textbf{MP: }(P $\to$ R) $\land$ P $\vdash$ R \\ \textbf{MP: }(R $\to$ S) $\land$ R $\vdash$ S \end{tabular} & 
\begin{tabular}[c]{@{}l@{}}  (P $\to$ Q), \\(Q $\to$ R), (R $\to$ S) \end{tabular} & 
P & 
\textbf{S: \checkmark}
% Is S true? \textbf{: Yes} 

\\ \midrule

4 &
\begin{tabular}[c]{@{}l@{}} \textbf{CD: }(P $\to$ Q) $\land$ (R $\to$ S) $\land$ (P $\lor$ R) $\vdash$ (Q $\lor$ S) \\ \textbf{DS: }(Q $\lor$ S) $\land$ $\lnot$Q $\vdash$ S \\ \textbf{MP: }(S $\to$ T) $\land$ S $\vdash$ T \\ \textbf{MP: }(T $\to$ U) $\land$ T $\vdash$ U \end{tabular} & 
\begin{tabular}[c]{@{}l@{}} (P $\to$ Q), \\ (R $\to$ S),  (P $\lor$ R), \\ (S $\to$ T), (T $\to$ U) \end{tabular} & 
$\lnot$Q & 
% Is U true? \textbf{: Yes}
\textbf{U: \checkmark}

\\ \midrule

5 &
\begin{tabular}[c]{@{}l@{}}\textbf{HS: }(P $\to$ Q) $\land$ (Q $\to$ R) $\vdash$ (P $\to$ R) \\ \textbf{MT: }(P $\to$ R) $\land$ $\lnot$R $\vdash$ $\lnot$P \\ \textbf{DS: }(P $\lor$ S) $\land$ $\lnot$P $\vdash$ S \\ \textbf{MP: }(S $\to$ T) $\land$ S $\vdash$ T \\  \textbf{MP: }(T $\to$ U) $\land$ T $\vdash$ U \end{tabular} &
\begin{tabular}[c]{@{}l@{}}(P $\to$ Q),\\ (Q $\to$ R),  (P $\lor$ S), \\(S $\to$ T), (T $\to$ U)\end{tabular}   & 
$\lnot$R & 
% Is U true? \textbf{: Yes}
\textbf{U: \checkmark}
\\
\bottomrule
\end{tabular}
}
\caption{Examples of multi-step reasoning rule combinations for PL. Similar combinations are used for FOL.}
\label{tab:examples_classical_logic_combinations}
\end{table*}
\subsection{Data Generation for Monotonic Logic}

Here, we provide details of the data generation process for PL and FOL (further details are in Appendix \ref{app:Monotonic_Logic}). Specifically, we delve into 14 distinct inference rules of PL and FOL, detailed in Table \ref{tab:classical_inference_rules}.

\paragraph{Choice of inference rules}
\label{sec 3:para:choice of inference rules}
Since entailment (concluding a formula in logic from another formula in that logic) in PL is Co-NP Complete, and entailment in FOL is undecidable. Even though we are interested in multi-step reasoning, our aim is not to build a ``complete'' reasoning system (the system that can make all possible entailments in that logic), rather, our goal is to make LLMs be able to at least mimic some key inference rules up to a depth of five, which itself is challenging. Thus, we start with the set of 25 inference rules used in \cite{parmar2024logicbench} and add eight more inference rules, resulting in 33 inference rules (with zero or one variable). For a depth of five that would mean a $33^5$ possible combination, which is already quite big ($>39$ million).
In addition, we also consider seven FOL inference rules involving three variables and binary, ternary relations (Appendix \ref{app:multi-variable fol}). In adding the new inference rules, our main consideration was how well they match human intuition. For example, we left out $p \wedge \neg p \vdash q$ as that is not very intuitive to non-logician humans. Similarly, we left out inference rules such as simplification $((p \land q) \vdash p)$, conjunction $(p, q \vdash (p \land q))$, and addition $(p \vdash (p \lor q))$, as they would lead to infinite reasoning chains and it did not make sense to add them as an additional step of reasoning to arrive at a meaningful conclusion. Conversely, we added the DMT ($\neg (p\land q)\dashv\vdash \neg p\lor \neg q$), and show its use in multi-step, as shown in Table \ref{tab:three_step_reasoning_combinations} (Appendix \ref{app:Rule_Combinations}).

\begin{table*}[ht]
\centering
\resizebox{0.95\linewidth}{!}{
\begin{tabular}{m{5.2cm}|m{14.2cm}}
\toprule
\multicolumn{1}{c|}{Rule Combination}    & \multicolumn{1}{c}{Context and Question} \\ \midrule \midrule

% % \textbf{Propositional Logic} \newline
\textbf{PL Rules:} MT, DS  \newline \textbf{Propositions:} \newline p: Capture shots in golden hours.\newline q: Photo wins awards.\newline r:  Focus on rare wildlife. & 
\textbf{Context:} In wildlife photography, Olivia was certain that if she captured shots in the golden hours, her photos would win awards. However, opportunities varied each day. It was evident that she either captured shots during the golden hours or she focused on rare wildlife, or both. Olivia's latest photos did not win any awards. \newline
% \newline 
\textbf{Question:}  Is it true that she focused on rare wildlife?\\ \midrule

% \textbf{First-order Logic} \newline 
\textbf{FOL Rules:} BD, DS 
\newline 
\textbf{Predicates:} \newline p: Work extra hours.\newline q: Meet project deadlines.\newline r: Take minimal breaks.\newline s: Increase productivity.
& 
\textbf{Context:} In a company, employees believe that if they work extra hours, they will meet project deadlines, and if they take minimal breaks, they will increase productivity. However, they face a dilemma - they either work extra hours or do not increase productivity. 
\newline 
% \newline
\textbf{Question:} Jane didn't meet the project deadline. Is it true that Jane took minimal breaks? 
\\ \midrule

% \textbf{Non-monotonic Logic} \newline 
\textbf{NM rule:} BDR \newline
\textbf{PL rule:} MP (Sentence Y)
\newline
\textbf{Logic:} Conclusion of BDR: \textit{X} \newline MP: (X $\to$ Y) $\land$ X $\vdash$ Y
& 
\textbf{Context:} Jim and Pam work at the same office. Normally, employees at that office get free lunch. Jim does not get free lunch. If Pam gets free lunch, then she gets an hour lunch break.
\newline 
% \newline
\textbf{Question:} Can we conclude Pam gets an hour lunch break?\\ 

\bottomrule

\end{tabular}
}
\caption{NL examples of different rule combinations for all three logic types. Appendix \ref{app:NL_examples} provides more examples.}
\label{tab:depth_examples_NL}
\end{table*}

\subsubsection{Generation of Rule Combination}
\label{sec:rule_comb_gen}
% \textcolor{red}{Aashna: will add an examples table for this}

% For introducing multi-step logical reasoning, we combine different inference rules that build upon each other, leading to a final conclusion. To achieve this, we chose combinations of rules that have a common premise or conclusion. In the combinations we choose, a rule's conclusion will be part of the premise of another rule. An illustration of this is shown in the figure \ref{fig:classical-logic-rule-combinations}.

% To incorporate multi-step logical reasoning into \textit{Multi-LogiEval}, we employ various inference rules that sequentially contribute to reaching a final conclusion as illustrated in Figure \ref{fig:classical-logic-rule-combinations}. 
%Our approach involves selecting rule combinations with shared premises or conclusions, which means that the conclusion of one rule becomes part of the premise for another rule in the chosen combinations. This process is illustrated in Figure \ref{fig:classical-logic-rule-combinations}.

\begin{figure}[ht]
    \centering
    \includegraphics[width=0.8\linewidth]{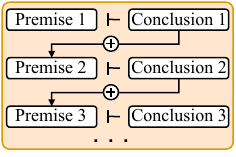}
    \caption{Process for combining multiple logical inference rules for PL and FOL: \textit{Premise 1} is the set of premises for the first inference rule, leading to \textit{Conclusion 1}. \textit{Conclusion 1} and \textit{Premise 2} derive \textit{Conclusion 2}, and so on. $\vdash$: Entails.}
    \label{fig:classical-logic-rule-combinations}
\end{figure}

% In this way, we ensure that a chain of reasoning will be required to answer the question. Some knowledge required to answer the question is given in the context, and the rest of the knowledge is provided in the question. The question needs to be answered based on knowledge from both the context and the question. Using this method, we generate combinations of varying steps. We consider 1 step to be using 1 rule from the basic inference rules to generate the context and question. To choose the combination, we first take the step 1 rule and see if any other rule contains a part of the premise or conclusion of this rule. This will form the 2-step combination, the conclusion of the current step becoming input into the next step.
We apply sequential inference rules for multi-step reasoning, as illustrated in Figure \ref{fig:classical-logic-rule-combinations}. To ensure a comprehensive approach to answering a question, we employ a method that involves leveraging contextual information and explicit details provided in the question. This process requires a logical chain of reasoning, combining knowledge from the given context with the information presented in the question. Each step in the reasoning chain corresponds to an inference rule, with combinations ensuring each step aligns with a single rule. To generate the combinations, we start with the initial rule and assess whether the conclusion of this rule aligns with the premise of other rules. This iterative process results in multi-step combinations/reasoning, with the conclusion of each step serving as a part of the premise for the subsequent rule. 
%to answer the question.
% \input{tables/classical_example_combinations}
% We consider the double negation inference rule a parallel step (negation of (``it is not raining”) is equivalent to ``it is raining”).
% With this in mind, we generated 25 combinations of rules, including 2-step, 3-step, 4-step, and 5-step. Table \ref{tab:examples_classical_logic_combinations} shows one example of a rule combination for each depth. To illustrate this, consider the combination of rules Modus Tollens ($((p \to q) \land \neg q) \vdash \neg p$) and Disjunctive syllogism ($((p \lor r) \land \neg p) \vdash r$) the context given in the story will be $(p \to q)$, $(p \lor r)$ and in the question additional context provided is $\neg q$. We then ask if $r$ is true. To answer this, first, from the defined Modus Tollens inference rule using the given context $(p \to q)$ from the story and $\neg q$ from the question, we can conclude $\neg p$, and using this derived $\neg p$ as the premise to Disjunctive Syllogism inference rule, $(p \lor r)$ from the story, we can conclude that $r$ is indeed true. A natural language illustration is shown in Table \ref{tab:NL_example_pl}. All our generated combinations can be found in the appendix.
% \textcolor{red}{a natural language example table.

%To explore various scenarios, 

We create 71 rule combinations, ranging from 2-step to 5-step reasoning chains. We use each single inference rule as depth-1. Examples of rule combinations in classical logic are presented in Table \ref{tab:examples_classical_logic_combinations}. Let's consider a specific combination involving the \textit{Modus Tollens} ($((p \to q) \land \neg q) \vdash \neg p$) and \textit{Disjunctive Syllogism} ($((p \lor r) \land \neg p) \vdash r$) rules for creating combination for depth-2. Given the context, including natural language statements for $(p \to q)$ and $(p \lor r)$ and information in the question as $\neg q$, we ask about the truth value of $r$. Applying \textit{Modus Tollens}, we deduce $\neg p$ from the $(p \to q)$ from context and $\neg q$ in question, giving the first step. Subsequently, using $\neg p$ as the premise for \textit{Disjunctive Syllogism}, we conclude that $r$ is indeed true based on the $(p \lor r)$ and $\neg p$, giving the second step. Creating rule combinations at higher depths, and validating the quality of generated instances is challenging, hence, we limit the number of rule combinations at $d_5$ for the scope of this study. More examples provided in Appendix \ref{app:Rule_Combinations}.

%We provide natural language examples in Table \ref{tab:depth_examples_NL}.
% Since most inference rules contain a disjunction, naturally, Disjunctive Syllogism became part of most of our rule combinations. 
% \textcolor{red}{Aashna: will add all full tables in appendix}.

% \begin{itemize}
% \item We consider 1 step to be using 1 rule to generate an output.
%         \item Output of the current step becomes an input into the next step.
%         \begin{itemize}
%             \item Example: (Can add a figure for this)
%         \end{itemize}
%         \item We consider double negation to be a parallel step.
%         \item We generated 25 combinations including 2-step, 3-step, 4-step and 5-step combinations.
%         \begin{itemize}
%             \item (How did we choose these 25?)
%         \end{itemize}
%         \item All our generated combinations can be found in the figure/appendix.
% \end{itemize}

\subsubsection{Generation of Data Instances}
% \textcolor{red}{Aashna: will add a table for data generation example of prompt+response.}

% To generate the data, we prompt the Claude 2 model with instructions tailored to each rule combination. We used a few-shot prompting technique to generate the data. Our data creation prompt structure, as illustrated in Figure \ref{fig:classical-logic-prompt-structure}, for both Propositional and First-Order Logic, is divided into the following subsections where we explain the relevance of each prompt subsection.
%\footnote{\url{https://www.anthropic.com/index/claude-2}}
% To create natural language (NL) data instances corresponding to various depths, we prompt Claude-2 with instructions corresponding to various rule combinations. To enhance the data generation process, we utilize a few-shot prompting. 
We generate natural language (NL) data at different depths by prompting Claude-2 in a few-shot setting with instructions for various rule combinations. The prompt schema, shown in Figure \ref{fig:classical-logic-prompt-structure}, comprise five crucial components:  

\begin{figure}
    \centering
    \includegraphics[width=0.75\linewidth]{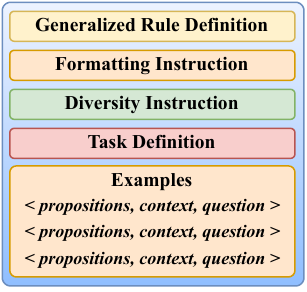}
    \caption{Data generation prompt for PL and FOL}
    \label{fig:classical-logic-prompt-structure}
\end{figure}

% An Example of a prompt created for depth 2 is presented in the Appendix.

\paragraph{Rule Definition} 
We provide generalized rules for various combinations containing propositions represented by labels such as {P} and {Q}. For instance, Rule 1: ``If {P} is true, then {Q} is true.'' Utilizing these defined rules, we construct the contextual premise by combining them. Subsequently, we formulate a question that requires a step-by-step deduction using all the established rules to derive the answer. 

%This structured approach allows for a comprehensive exploration of knowledge within the given context. 

% For each combination of rules, we first define the set of generalized rules for the currently chosen combination to represent the knowledge given in the context.  We use labels \{P\}, \{Q\}, … to denote propositions or predicates. An example of rule definition: \textit{Rule 1: [if \{P\} is true, then \{Q\} is true.]} Using this set of rules, we then generate the context that forms the premise of the question.  We then define the question, which will necessitate using all the rules step by step to deduct the answer.

\paragraph{Format} We provide model-specific instructions for generating outputs in a designated format, simplifying the process of parsing it on a large scale. 

%For instance, we instruct the model with a formatting instruction like, "Complete the following tasks, only returning text in exactly the format given in the following examples. Do not generate example numbers when formatting the output." This approach streamlines the handling and analysis of generated content, enhancing efficiency in large-scale applications.
% We provide the model instructions to generate outputs in a particular format to facilitate parsing the output at a large scale. Example: \textit{Complete the following tasks, only returning text in exactly the format given in the following examples. Do not generate example numbers when formatting the output.

\paragraph{Diversity} 
To enhance diversity, we prompt the model to generate multiple instances across various domains, such as education and finance.

%in generated examples
%For instance, by instructing the model to generate 10 examples spanning different domains, we ensure a broader and more varied set of outputs. This approach contributes to the overall quality and richness of the generated content.
% Generating multiple examples at the same by explicitly prompting to use different domains ensures that diversity is introduced. Example: \textit{Generate 10 more examples from multiple domains.}

% \input{tables/NM_exampls}

\paragraph{Task Definitions} We provide definitions to perform two tasks. First, to generate the context that serves as a human-like illustration of generalized rules. This task instructs the generation of a real-life story with sentences exemplifying the specified rules, where entity labels such as $P, Q, R, S, T,$ and $U$ are replaced with actual entities. To ensure clarity, entity labels are excluded from the context. Additionally, the context generation for FOL incorporates instructions specifying the use of generalized sentences with indefinite pronouns for quantification. The second task focuses on question generation, which entails formulating questions in the format: "[(....) is true/not true, then is (....) true?]" This dual-task approach ensures the generation of \textit{<context, question>} pair. We provide examples of generated NL instances in Table \ref{tab:depth_examples_NL}.
%for PL and FOL.
% We define two tasks for the data generation process: one for generating the context story and another for generating the question. The story generation task is an instruction to create a human-like story to illustrate the generalized rules.  We also give an instruction to replace the entity labels \{P\}, \{Q\}, \{R\} … in the generalized rules with actual entities. An example of a story generation task:  Generate a short real-life story that includes sentences to illustrate the above rules, replacing the entities P, Q, R, S, T, and U with real entities. Do not include the entity labels like P, Q, R, S, T, and U in the story. First-order logic has additional instructions only to use generalized sentences using indefinite pronouns to resemble quantification. The question generation task contains the definition of a question of the form: [If (....) is true/not true, then is (....) true?]. 

\paragraph{Examples}
We present five in-context exemplars for every rule combination. Each instance comprises propositions such as $P, Q, R$, a contextual narrative, and an associated question. An example prompt for depth-3 is presented in Appendix \ref{app:prompt}, and we use a similar structure for all other prompts.

\subsection{Non-Monotonic Reasoning}
\label{subsec:nm_data}
% Here, we utilize eight NM reasoning patterns defined in the \citet{lifschitz1989benchmark}, provided in Appendix \ref{app:NM}. For NM, we only generated data for depth-1 and depth-2 of logical difficulty. We limit our data generation to depth-2 since the NM reasoning patterns presented in \citet{lifschitz1989benchmark} involve 4-5 assumptions for each rule, and combining two rules with classical logic results in a lengthy narrative and it becomes challenging for LLMs to generate quality instance with that long narrative. Hence, we limit NM to depth 2.  
We utilize eight NM reasoning patterns defined in \citet{lifschitz1989benchmark} (Appendix \ref{app:NM}), and have generated data for depths 1 to 5. To increase reasoning depth, we integrated NM with classical logic, using only one NM rule per depth due to the 4-5 assumptions each pattern involves. Thus, combining two NM patterns with classical logic creates lengthy contexts, challenging for LLMs to generate quality instances. Our rule combinations avoid overly long contexts while requiring reasoning up to depth-5.

%Thus, even combining two NM patterns with classical logic results in a lengthy context, making it challenging for LLMs to generate quality instances. We combine rules not to include overly long contexts but still require up to five depths of reasoning.
%presented in \citet{lifschitz1989benchmark} with such extended narratives
%For depth 2, we combined the inference rules from PL with NM reasoning patterns. Examples are provided in Table \ref{tab:NM_NL_Example_D2} (Appendix \ref{app:NM}).
%Moreover, exploring deeper depths in non-monotonic reasoning becomes impractical, as the model requires at least 8-9 reasoning steps to reach accurate conclusions. This is comparable to the depth-4 or 5 in classical logic, and further depth is unnecessary. 
\paragraph{Generation of Rule Combination}
We consider reasoning patterns corresponding to default reasoning for depth-1. We generalize the rule to generate simple sentence pairs independently before combining them according to the template-based NM rule. After generating sentence pairs, we combined the sentences based on the defined rule and formulated the question-answer pair accordingly. We have manually generated 12, 2, 2, and 1 rule combinations for depth-2, depth-3, depth-4, and depth-5, provided in Appendix \ref{app:NM}.
While formulating depthwise rule combinations, a logical relationship between the context and question is followed. The rule combinations for all depths from 2 to 5 include 6 reasoning rules from NM—BDR, PBD, DRO, PBD, REII, and REIII—and 3 inference rules from PL—MP, MT, and DS. The data for depths 2 to 5 is generated by forming a logical connection between two NM rules' conclusions and the PL rules.

%% We consider reasoning patterns corresponding to default reasoning for depth-1. We generalize the rule to generate simple sentence pairs independently before combining the template-based NM rule. After generating sentence pairs independently, we combined the sentences based on the defined rule and formulated the question-answer pair accordingly. To achieve the rules with reasoning depth-2, we combined the rules from PL and NM. We manually generate a total of 9 such rule combinations provided in Appendix \ref{app:NM}. A logical relationship between context and question is followed while formulating depth-2 rule combinations. The rule combinations include 7 rules from NM - BDR, DRI, PBD, DRO, PBD, REII, and REIII and 3 inference rules from PL - MP, MT, and DS. The overall depth-2 data is generated by establishing a logical connection between the conclusions of two NM patterns with the PL rules. 
%The conclusion of two non-monotonic rules logically follows the defined propositional rule for all 9 different combinations.

% \input{tables/NM_2_example}

\paragraph{Generation of Data Instances} 
In creating prompts for data generation, we use a four-part structure: (1) define the task, (2) explain each rule as an assumption and conclusion, (3) provide instructions for creating context and questions to ensure logical connections, and (4) establish formatting guidelines for systematic output. Appendix \ref{app:NM} shows an example of the prompt.
% In creating prompts for data generation, we adopt a four-part structure. First, we define the task. Second, each rule is explained in the prompt, representing a broad assumption and conclusion. 
% Third, we give clear instructions for creating context and questions. These instructions guide the model's adherence to specific rules, ensuring a logical connection. Finally, we establish formatting guidelines, ensuring a systematic model output. Appendix \ref{app:NM} shows an example of the prompt.

%our method employs instruction-based generation, where 
% We use $X$ to signify the conclusion of rule 1 and $Y$ for rule 2.
% To create data generation prompts, we adopt a four-part structure. First, we provide the task definition. Second, each of the three rules is explained in the prompt, encapsulating a generalized assumption and conclusion, denoting $X$ as the conclusion of rule 1 and in the same manner $Y$ for rule 2. Thirdly, our approach involves instruction-based generation, where we give clear and detailed instructions to create context and questions. These instructions guide the model to follow specific rules for generating context and corresponding questions, ensuring a logical connection in the process. Finally, we provide formatting guidelines for generation, ensuring a systematic output from the model. An example of a prompt is provided in Appendix \ref{app:NM}.

\subsection{Qualitative Analysis}
\label{sec:qualitative_analysis}

% After data generation, we conducted a manual qualitative analysis before including the sample in the \textit{Multi-LogicEval}. This process resulted in high-quality $1552$ samples spanning three different logic types and various reasoning depths.
After data generation, we conducted a manual qualitative analysis, resulting in 1,552 high-quality samples for \textit{Multi-LogicEval}.

\begin{table}[!hbpt]
\centering
\resizebox{\linewidth}{!}{
\begin{tabular}{l|c|c|c|c|c|c}
\toprule
\multicolumn{1}{c|}{\multirow{2}{*}{\textbf{Logic}}} & \multicolumn{5}{c}{\textbf{Reasoning Depth}} & \multicolumn{1}{|c}{\multirow{2}{*}{\textbf{Total}}} \\ \cmidrule{2-6}
\multicolumn{1}{c|}{}                          & 1 & 2   & 3   & 4   & 5  & \\ \midrule\midrule
PL                          & 120  & 105 & 135 & 120 & 45 & 525\\
FOL                          & 130  & 105 & 135 & 120 & 45 & 535\\
NM                           &  160 & 232 & 40   & 40   & 20   & 492\\ \midrule\midrule
\textbf{Total}                         &  410 & 442 & 310 & 280 & 110 &  \textbf{1552} \\ \bottomrule
\end{tabular}
}
\caption{Statistics of \textit{Multi-LogiEval}}
% : Number of samples for each depth.}
\label{tab:stats}
\end{table}

\paragraph{Statistics} 
\textit{Multi-LogicEval} has 5 different logical reasoning depths. Table \ref{tab:stats} shows the depth-wise statistics of samples present for each logic type after validation. After manual validation, from the generated data, we selected/updated high-quality 10 data instances for each inference rule in depth 1 and 15 or 20 data instances for each rule combination, which resulted in 410, 442, 310, 280, and 110 samples for depth-1, depth-2, depth-3, depth-4, and depth-5, respectively. For evaluation, of the total 1552 samples, 1126 samples have the answer \textit{yes}, and the remaining 426 samples have the answer \textit{no}.

%We have in total 1655, 1360, and 609 samples for PL, FOL, and NM for evaluation.

%Initially, we generate 50 samples corresponding to each rule combination. 

%The validation questions for non-monotonic rules(depth-2) are created in a way that validates the logical connection between rules in the generated context and ensures that the conclusion of both non-monotonic rules logically follows the mentioned propositional rule. The validation questions ensure the quality and logical connection between context and questions. The questions ensure that the context and question are not generalized, refer to specific objects and/or properties, language, and follow the logical structure mentioned in the generating samples. 

%\textit{Multi-LogicEval} has 5 different logical reasoning depths. Table \ref{tab:stats} shows the depth-wise statistics of samples present for each logic type after validation. After manual validation, from the generated data, we selected/updated high-quality 10 data instances for each inference rule in depth 1 and 15 or 20 data instances for each rule combination, which resulted in 410, 442, 310, 280, and 110 samples for depth-1, depth-2, depth-3, depth-4, and depth-5, respectively. For evaluation, of the total 1552 samples, 1126 samples have the answer \textit{yes}, and the remaining 426 samples have the answer \textit{no}.

% d1
% 73/410
% 17.8% 
% d2
% 93/442
% 21% 
% d3
% 73/310
% 23.5%
% d4
% 93/280
% 33.2%
% d5
% 23/110
% 21%

% pl - 115/525 ~22%
% fol - 102/535 ~19%
% nm - 127/492 ~25.8%

\paragraph{Quality of Data Instances} We examine each context for potential discrepancies throughout the data generation phase, ensuring they are logically correct and represent the intended logical relations. We also dedicated considerable effort to eliminating typos and validating the grammar. While validating, we encountered a few errors within the synthetically generated story-based context. We manually mitigate these errors to ensure integrity and utility (Analysis presented in Appendix \ref{app:Data_Validation}). 

\begin{table*}
\centering
\resizebox{\linewidth}{!}{
\begin{tabular}{c|c|c|c|c|c|c|c|c|c|c|c|c|c|c|c}
\toprule
\multirow{2}{*}{\textbf{Models}} & \multicolumn{5}{c|}{\textbf{Propositional}}                                                                                                             & \multicolumn{5}{c|}{\textbf{First-Order}}                                                                                                               & \multicolumn{5}{c}{\textbf{Non-Monotonic}}    \\ \cmidrule{2-16} 

& \textbf{$d_1$} & \textbf{$d_2$} & \textbf{$d_3$} & \textbf{$d_4$} & \textbf{$d_5$} & \textbf{$d_1$} & \textbf{$d_2$} & \textbf{$d_3$} & \textbf{$d_4$} & \textbf{$d_5$} & \textbf{$d_1$} & \textbf{$d_2$} & \textbf{$d_3$} & \textbf{$d_4$} & \textbf{$d_5$} \\ \midrule
GPT-4                            & 89.17      & 69.52      & 82.22      & 71.67      & 66.67       & 83.85      & 70.48      & 71.85      & 59.17      & 66.67       & 36.88      & 51.67   &   65.00	& 67.50	& 60.00    \\ 
ChatGPT                          & 91.67      & 56.19      & 63.70      & 62.50      & 44.44       & 97.69      & 59.05      & 57.78      & 50.83      & 37.78       & 33.75      & 41.11    &  50.00 & 	62.50 &	60.00 \\ 
Gemini                           & 90.00      & 62.86      & 68.15      & 65.83      & 60.00       & 76.92      & 62.86      & 65.93      & 57.50      & 53.33       & 46.25      & 46.11    &  62.50	& 55.00 &	60.00 \\ 
Yi-34B                           & 85.00      & 65.71      & 58.52      & 46.67      & 26.67       & 90.00      & 55.24      & 57.94      & 48.33      & 13.33       & 37.50      & 41.11     & 55.00	& 62.50 &	65.00 \\ 
Orca-13B                         & 75.83      & 41.91      & 35.56      & 35.00      & 15.56       & 66.92      & 47.62      & 42.96      & 40.00      & 6.67        & 21.88      & 26.67    &  25.00 &	15.00 &	25.00 \\
Mistral-7B                       & 80.83      & 68.57      & 61.48      & 53.33      & 44.44       & 83.85      & 63.81      & 56.30      & 52.50      & 20.00       & 37.50      & 39.44    &  52.50	& 47.50 &	65.00  \\ \midrule
\textbf{Avg}                     & \textbf{85.42}      & \textbf{60.79}      & \textbf{61.61}      & \textbf{55.83}      & \textbf{42.96}      & \textbf{83.21}      & \textbf{59.84}      & \textbf{58.79}      & \textbf{51.39}      & \textbf{32.96}      & \textbf{35.63}      & \textbf{41.02}  & \textbf{51.67}  & \textbf{51.67} & \textbf{55.83} \\ \bottomrule
\end{tabular}
}
\caption{Evaluation of LLMs in terms of accuracy on \textit{Multi-LogiEval}.}
\label{tab:main_results}
\end{table*}

\section{Results and Analysis}
\label{sec:results}

% In this section, we present detailed information related to the experimental setup, our primary results, and a detailed analysis of the results.

\subsection{Experimental Setup}

\paragraph{Task Formulation}
We formulate a binary classification task using \textit{Multi-LogiEval}.
Let us consider a set of data instances $\mathcal{I}_{D, L}$ corresponding to depth $D$ and logic type $L$. In this set, $i^{th}$ instance is represented as $\mathcal{I}^i_{D, L} = \{(c_i, q_i)\}$ where $c_i$ represents context and $q_i$ represents question corresponding to $i^{th}$ instance. Here, each context and question pair is created so that the conclusion provided in the question always entails context. However, you require different reasoning steps to conclude. We prompt the model to assign a label \textit{Yes} if the conclusion logically entails the context; otherwise, \textit{No}. To evaluate any LLMs, we provide $<p,c,q>$ as input to predict a label \textit{Yes} or \textit{No} where $p$ is a natural language prompt. 

%Each context ($c$) represents a story embedded with natural language logical statements, and question ($q$) represents the conclusion (see Table \ref{tab:examples_classical_logic_combinations}).
%to evaluate LLMs' multi-step logical reasoning ability , \textit{Multi-LogiEval} consists of data instances with different reasoning steps/depths.
%In this work, we use a zero-shot chain of thought to evaluate LLMs.

\paragraph{Experiments}
We evaluate a range of proprietary models (i.e., GPT-4, ChatGPT, and Gemini-Pro) and open-source models (i.e., Yi-34B-Chat, Orca-2-13B, and Mistral-7B-Instruct) on \textit{Multi-LogiEval}. The evaluation is conducted on the versions of OpenAI and Google models released in April 2024. Each model is evaluated in a zero-shot-CoT setting. The prompt used for experiments is provided below. We evaluate LLMs in a zero-shot setting to show the logical reasoning ability of the model based on knowledge acquired during pre-training since we can not expect in-context examples corresponding to different reasoning patterns and depths during inference. However, we also evaluate LLMs in a 3-shot setting (results are in Appendix \ref{app:fewshot_val}).

%where the chain-of-thought prompt is provided to the model without any in-context examples

\paragraph{Metrics} Since the objective is to assess the model's ability to arrive at the correct conclusion, we measure the accuracy associated with a \textit{Yes} and \textit{No} label. Apart from accuracy, we provide an in-depth analysis of reasoning chains in section \ref{subsec 4.3: Analysis and Discussion} to gain insights into models' performance. In addition, we would like to mention that the binary labels \textit{Yes} and \textit{No} indicate whether the conclusion presented in the question can be derived from the context. Hence, accuracy is an important evaluation metric, reflecting the model's reasoning ability.

\begin{tcolorbox}[colbacktitle=black!80!black]
{\small Given the context that contains rules of logical reasoning in natural language and question, perform step-by-step reasoning to answer the question. Based on context and reasoning steps, answer the question ONLY in `yes' or `no.' Please use the below format:
\newline \textbf{Context:} [text with logical rules] \newline \textbf{Question:} [question that is based on context] \newline \textbf{Reasoning steps:} [generate step-by-step reasoning] \newline \textbf{Answer:} Yes/No
}
\end{tcolorbox}

%We evaluate performance in terms of accuracy. 
%based on the answer to a given question

\subsection{Main Results}

\paragraph{Objective Evaluation} Table \ref{tab:main_results} illustrates the accuracy of reasoning at different depths for various LLMs, offering significant insights into their performance across distinct logic types and depths.
From Table \ref{tab:main_results}, experimental results reveal a consistent trend across PL and FOL, i.e., as the reasoning depth increases from 1 to 5, the models' average performance drops. 
In particular, at depths 4 and 5, accuracy drops significantly for the majority of LLMs we evaluated. 
For instance, the accuracy of GPT-4, ChatGPT, and Gemini demonstrates a substantial drop from $89.17\%$, $91.67\%$, and $90\%$ at $d_1$ to $66.67\%$, $44.44\%$, and $60.00\%$ at $d_5$ for PL, respectively, indicating the challenge encountered even by larger-scale LLMs when handling longer chains of logical reasoning. 
In summary, for PL and FOL, LLMs perform well on $d_1$ compared to other depths. While they show competitive performance for $d_2$ and $d_3$, there is a significant drop in performance for $d_4$ and $d_5$ in most cases.
In contrast, moving on to NM, going from $d_1$ to $d_5$, there is an increase in the performance of LLMs from an average of $35.63\%$ to $55.83\%$. 

\paragraph{Random Baseline} We calculated a random baseline for each depth from Multi-LogiEval as below:

\[
\text{Acc}_{\text{random}} = p_{\text{yes}}^2 + p_{\text{no}}^2,
\]

where $p_{\text{yes}}$ and $p_{\text{no}}$ represent the probabilities of predicting ``yes'' and ``no,'' respectively. The random baselines for depths $d_1$, $d_2$, $d_3$, $d_4$, and $d_5$, with corresponding $\text{Acc}_{\text{random}}$ of 86.63\%, 67.35\%, 53.71\%, 58.33\%, and 83.33\%, respectively. From Table \ref{tab:main_results}, we can observe that these models perform lower in terms of average accuracy compared to the random baseline.

\paragraph{Findings} Table \ref{tab:main_results} reveal that open-source models experience a significant performance drop from $d_4$ to $d_5$. Also, there is an increasing performance trend in NM. For PL and FOL, GPT-4, ChatGPT, and Gemini show improved performance from $d_2$ to $d_3$, whereas the performance of open-source models consistently decreases. In addition, larger open-source models demonstrate decreasing performance. Furthermore, ChatGPT performs lower than GPT-4 and Gemini at $d_5$ in PL and FOL. Also, FOL performance is lower compared to PL at $d_5$.

\subsection{Analysis and Discussion}
\label{subsec 4.3: Analysis and Discussion}
In this section, we manually analyze the generated reasoning chains \footnote{\url{https://github.com/Mihir3009/Multi-LogiEval}} by different LLMs and investigate the above-mentioned findings in detail. 

%\url{https://anonymous.4open.science/r/Multi-LogiEval-FFDB}
% \paragraph{Depth \textit{vs.} Performance Drop} The relationship between model depth and performance is a critical aspect of natural language processing. In the investigation of "Depth vs. Performance Drop," it was observed that as the depth of reasoning increases, there is a notable decline in models' performance, particularly evident in depths 4 and 5, posing significant challenges. Notably, models with depths 1, 2, and 3 exhibited comparable performance, while a substantial decrease was observed with greater depths 4 and 5. The increase in context length across all models was identified as a contributing factor to the performance drop. Notably, GPT-3 outperformed other models in predictive tasks, such as Programming Language (PL) and First-Order Logic (FOL), highlighting the inherent difficulty that arises as model depth increases in making accurate predictions.

\paragraph{Performance Improvement from $d_2$ to $d_3$ in PL and FOL for GPT-4, ChatGPT, and Gemini}
% Large-scale models such as GPT-4, ChatGPT, and Gemini demonstrate higher performance at $d_3$ for PL, with a performance decrease as the depth increase ($d_4$ and $d_5$). This trend is also observed in FOL for the same models except ChatGPT. A systematic examination of reasoning chains for PL and FOL for these models revealed that larger models often perform better at $d_3$ compared to $d_2$, often due to the models reaching incorrect conclusions from wrong interpretation of evidence. 
% In $d_3$, increasing context length makes these LLMs more accurate in mapping information, thus achieving peak performance (comparison with $d_2$ to $d_5$). We manually analyzed all the reasoning chains with wrong predictions for these models. We observe that at $d_2$, $\sim27.4\%$ reasoning chains with incorrect conclusions resulted from the models' inability to map information correctly, either the premise from the context to the correct conclusion or the premise from the previous step to the right conclusion. This observed number is dropped to $\sim22\%$ at $d_3$ and we observed that a larger context length at $d_3$ helps in reducing this problem. However, at $d_4$ and $d_5$, the length of the reasoning chain increases. Longer reasoning steps are more prone to error propagation at later stages, leading to incorrect conclusions and lower performance. Additionally, as the context becomes longer and more multi-step, the number of possible reasoning paths increases, causing the models to deviate further from the true conclusion. 
GPT-4, ChatGPT, and Gemini excel at $d_3$ for PL, with a performance decrease at $d_4$ and $d_5$. This trend is also observed in FOL for the same models except ChatGPT. Systematic analysis of all the reasoning chains with wrong predictions for PL and FOL shows these models reach incorrect conclusions often due to the wrong interpretation of evidence. 
In $d_3$, increasing context length improves LLMs accuracy in information mapping, thus achieving peak performance (comparison with $d_2$ to $d_5$). 
At $d_2$, around $\sim27.4\%$ of reasoning chains with incorrect conclusions were due to the models' failure to correctly map information, either from context to conclusion or the premise from one step to the next step. This number drops to $\sim22\%$ at $d_3$ and we observed that a larger context length at $d_3$ helps in reducing this problem. However, at $d_4$ and $d_5$, the length of the reasoning chain increases further. Since longer reasoning steps are more prone to error propagation at later stages, causing the models to deviate further from the true conclusion, hence, lower performance at $d_4$ and $d_5$.

\paragraph{Lower Performance of ChatGPT compared to GPT-4 and Gemini at Higher Depths}
% This pattern is particularly evident in FOL and PL at $d_5$ for ChatGPT compared to Gemini, and GPT-4.
% At d5, by analyzing the reasoning chains manually, we observe that ChatGPT often generates longer reasoning chains compared to Gemini, and GPT-4 to answer the same question. Specifically, for PL and FOL, the average length of the reasoning chain for ChatGPT at $d_5$ is 13.85. In contrast, the average length of the reasoning chain for Gemini and GPT-4 at $d_5$ is 8.85 and 10.87, respectively. This highlights that longer reasoning chains do not necessarily correlate with better reasoning outcomes since they are more prone to error propagation at later stages. This observation indicates the complexity of reasoning tasks at higher depths and suggests that optimizing the length of the reasoning chain is crucial for enhancing model accuracy in complex scenarios.
This pattern is particularly evident in FOL and PL at $d_5$ for ChatGPT compared to Gemini, and GPT-4.
At d5, manual analysis shows that ChatGPT tends to generate longer reasoning chains compared to Gemini, and GPT-4 when answering question. For PL and FOL, the average reasoning chain length for ChatGPT at $d_5$ is 13.85, while for Gemini and GPT-4 at $d_5$ is 8.85 and 10.87, respectively. Longer reasoning chains do not necessarily correlate with better reasoning outcomes, highlighting the complexity of complex reasoning task. This suggests that optimizing reasoning chain length is crucial for improving model accuracy in complex scenarios.

\paragraph{Increasing Performance Trend in NM}
% We observe that the models demonstrate consistent patterns of improving performance as the depth increases for NM. Specifically, we analyze the reasoning chain systematically for ChatGPT and the open-source model (Yi-34B) where this trend is prominent. This divergent behavior compared to classical logic (PL and FOL) can be attributed to rule combinations created using NM reasoning patterns for $d_2$ to $d_5$. As discussed in section \ref{subsec:nm_data}, there is only one NM reasoning pattern present for these depths, and achieving multi-step reasoning by combining it with PL inference rules. For instance, $d_2$ combines 1 PL rule with 1 NM reasoning pattern, $d_3$ uses 2 PL rules with 1 NM reasoning pattern, and so forth. By observing the reasoning chains and referring to the reason for the higher performance of models for PL and FOL at $d_3$, we conclude that the NM reasoning pattern present at each depth provides supplementary evidence and information, enhancing the models' contextual understanding. As the depth increases, adding one simple classical rule to the NM bolsters the models' accuracy in their ability to conclude. This is the primary reason for the significant performance improvement in NM compared to classical logic at the more challenging depths of 4 and 5.
In our analysis of ChatGPT and the open-source model Yi-34B, we've observed consistent performance improvements with increasing depth in NM reasoning. 
This trend diverges from classical logic PL and FOL. Specifically, at depths $d_2$ to $d_5$ , NM exhibits novel performance due to unique rule combinations in reasoning patterns. For instance, at $d_2$, NM combines one PL rule with one NM reasoning pattern, progressing to two PL rules with one NM pattern at $d_3$, and so forth. The addition of NM reasoning patterns complements PL and FOL by providing supplementary evidence and improving contextual understanding. Notably, as depth increases, integrating basic classical rules with NM significantly enhances model accuracy, particularly evident at depths 4 and 5. This integration is pivotal for the notable performance gains observed in NM compared to classical logic at higher depths.

\paragraph{Larger Open-Source Models Show Decreased Performance Compared to Smaller Models}
Here, we examine Mistral-7B, Orca-13B, and Yi-34B, which differ significantly in parameter size. Mistral-7B, the smallest, performed best across various depths of classical logic, except at the simplest $d_1$. As reasoning depth increased, Mistral-7B consistently outperformed Orca-13B and Yi-34B, with Yi-34B only marginally better ($1.5\%$) at $d_3$. For NM tasks, Mistral-7B and Yi-34B showed similar performance across all depths. At the most challenging depth ($d_5$) for both PL and FOL, Mistral-7B outperformed Orca-13B by achieving 3x performance despite Orca-13B's larger size. We believe that this capability of Mistral-7B is attributed to its architecture and training, enhancing its reasoning abilities, as discussed in \citet{jiang2023mistral}. In particular, the training of Mistral-7B focused on enhancing its reasoning capabilities.

\paragraph{Lower Average Performance in FOL than PL at $d_1$ to $d_5$} 
Upon observing the reasoning chains with wrong final predictions for the FOL and PL, we find that the generic rules in FOL contexts lead to deviations from the correct reasoning path. In some cases, it assigns predicates incorrectly to the FOL inference rule. This pattern is more prominent at $d_5$, highlighting the large gap ($\sim10\%$) in average performance between PL and FOL.
% In our analysis, models demonstrated a marked difficulty in reasoning at the most complex depth, d5, for both Propositional Logic (PL) and First-Order Logic (FOL). However, FOL consistently lagged behind PL in performance across all depths (d1-d5). Upon examining the reasoning chains, it became apparent that models struggle more with FOL due to the more generic context it has, leading to a higher likelihood of deviating from the correct reasoning path. In contrast, the specific cases presented in PL tasks facilitate more accurate predictions by the models.
% This disparity is particularly evident at the challenging depth d5, where the difference in performance between PL and FOL becomes significant. The generic nature of FOL makes it harder for models to map information accurately, whereas the specificity in PL scenarios aids in narrowing down the possible reasoning paths, resulting in better performance. This trend underscores the inherent challenges models face when dealing with the broader and more abstract contexts typical of FOL.

\paragraph{Lower Performance in $d_1$ of NM}
Reviewing the reasoning chains, we noted that models struggled to accurately map information. Interpreting various assumptions is crucial for effective reasoning at $d_1$. However, we observed that models have difficulty concluding based solely on assumptions present in the context when explicit knowledge is absent.
% Reviewing the reasoning chains, we noted that models encountered challenges in accurately mapping information. Interpreting various assumptions is crucial for reasoning effectively from the contexts of $d_1$. However, We observed that models struggle to conclude only based on assumptions present in the context and in the absence of explicit information.

\paragraph{Preliminary Discussion on Multi-variable FOL} 
Since our work focuses on evaluating LLMs' multi-step reasoning with simple FOL inference rules, we conducted only a preliminary study on their reasoning abilities for multi-variable FOL rules, discussed in Appendix \ref{app:multi-variable fol}. This study reveals that creating natural language instances is challenging for this kind of setup.

\paragraph{Case Study on Evaluating \textit{Multi-LogiEval} using Neural Symbolic}
Motivated by \citet{olausson-etal-2023-linc}, we evaluate GPT-4 using the neuro-symbolic approach where we utilized the Prover9\footnote{\url{https://www.cs.unm.edu/~mccune/prover9/}}. It first converts FOL statements into conjunctive normal form (CNF) and then performs resolution. Thus, we only evaluated data samples with FOL from \textit{Multi-LogiEval}. For this study, we adapt the evaluation approach presented in \citet{pan-etal-2023-logic} where GPT-4 is used to convert the context and question in a formal executable program and use Prover9 to solve it. We used an implementation with a similar GPT-4 version compatible with \citet{pan-etal-2023-logic}. We use the below prompt to convert natural language to an executable logic program:

\begin{tcolorbox}
Given a problem description and a question. The task is to parse the problem and the question into first-order logic formulas. The grammar of the first-order logic formula is defined as follows :
\\\\
1. logical conjunction: $expr1 \land expr2$\\
2. logical disjunction: $expr1 \lor expr2$\\
3. logical exclusive disjunction: $expr1 \oplus expr2$\\
4. logical negation: $\neg expr1$\\
5. $expr1$ implies $expr2$: $expr1 \rightarrow expr2$\\
6. $expr1$ if and only if $expr2$: $expr1 \leftrightarrow expr2$\\
7. logical universal quantification: $\forall x$\\
8. logical existential quantification: $\exists x$\\
\\
Output format: <logic form ::: description>
\end{tcolorbox}

We evaluate the performance in terms of the accuracy of selecting the correct answer. We also report the ``Executable Rate'', which reflects the grammar correctness of the logical form, and the ``Executable Accuracy'' of the executable samples to measure the semantic correctness \cite{pan-etal-2023-logic}.

\begin{table}[htbp]
    \centering
    \resizebox{\columnwidth}{!}{  % Resize to fit the column width
        \begin{tabular}{>{\centering\arraybackslash}m{1.8cm} >{\centering\arraybackslash}m{2.3cm} >{\centering\arraybackslash}m{2.8cm} >{\centering\arraybackslash}m{2.8cm}}
            \toprule
            \textbf{Reasoning Depth} & \textbf{Overall Acc (\%)} & \textbf{Executable Rate (\%)} & \textbf{Executable Acc (\%)} \\
            \midrule
            $d_1$ & 55.83 & 85.83 & 51.46 \\
            $d_2$ & 46.67 & 76.67 & 47.83 \\
            $d_3$ & 38.89 & 77.78 & 30.00 \\
            $d_4$ & 40.00 & 77.78 & 41.43 \\
            $d_5$ & 60.00 & 77.78 & 62.86 \\
            \bottomrule
        \end{tabular}
    }
    \caption{Performance Metrics by Reasoning Depths}
    \label{tab:neuro_sym}
\end{table}

As the reasoning depth increases from $d_1$ to $d_4$ (Table \ref{tab:neuro_sym}), there is a general trend of decreasing overall accuracy and executable accuracy, indicating that higher reasoning depth poses more challenges for executing code correctly. Interestingly, at $d_5$, both overall accuracy and executable accuracy show a significant improvement.

\paragraph{Human Evaluation and Further Discussion} 
In this study, we performed a human evaluation on a selected subset of the \textit{Multi-LogiEval}. Additionally, we explored potential strategies for enhancing LLMs' reasoning capabilities based on the findings of our current analysis. Please refer to Appendix \ref{app:human_eval} for further details on the human evaluation process and discussion.

\section{Conclusions}

In this work, we introduced \textit{Multi-LogiEval}, a comprehensive multi-step logical reasoning benchmark consisting of three types of logic and over 60 combinations of inference rules.  Our approach utilized two-stage methodology to construct data instances for our benchmark consisting of $\sim1.6k$ data instances with $1\sim5$ reasoning depth. We evaluated a range of LLMs, including GPT-4, ChatGPT, Gemini, Yi, Orca, and Mistral on \textit{Multi-LogiEval}. Experimental results revealed that these models struggle to perform logical reasoning, and their performance drops as the depth of logical reasoning increases (average accuracy of $\sim68\%$ at $d_1$ to $\sim43\%$ at $d_5$) for classical and non-classical logic. Furthermore, we systematically analyzed the reasoning chain generated by LLMs at various depths and presented interesting findings. We hope that \textit{Multi-LogiEval} will facilitate future research in evaluating and enhancing the ability of existing and upcoming LLMs for multi-step logical reasoning.

\section*{Limitations}
Though \textit{Multi-LogiEval} facilitates the evaluation of the multi-step logical reasoning ability of LLMs, the complexity of reasoning depth presented in \textit{Multi-LogiEval} can be improved by adding reasoning depth beyond five steps. \textit{Multi-LogiEval} can be further extended by incorporating other inference rules and logic types, for instance, the inference rules in first-order logic that capture n-ary relations between multiple variables. We also note that this research is limited to the English language and can be extended to multilingual scenarios for evaluating the logical reasoning ability of LLMs.

\section*{Ethics Statement}

We have used AI assistants (Grammarly and ChatGPT) to address the grammatical errors and rephrase the sentences.

\section*{Acknowledgement}
We thank the anonymous reviewers for their constructive suggestions and feedback. We extend our gratitude to the Research Computing (RC), and Enterprise Technology at ASU for providing computing resources, and access to the ChatGPT enterprise version for experiments. We acknowledge support by a 2023 Spring Amazon Research Award (ARA). This material is also based upon work supported by the Engineering Research and Development Center - Information Technology Laboratory (ERDC-ITL) under Contract No. W912HZ24C0022.

\bibliography{custom}

\clearpage

\appendix

\providecommand{\Mohith}[1]{
    {\protect\color{blue}{[Mohith: #1]}}
}
% \section{Extended Related Work}
\section{Monotonic Logic Description} \label{app:Monotonic_Logic}
\paragraph{Propositional Logic (PL)}
% Propositional logic uses propositions (sentences that can be categorized as true or false denoted as ${p, q, r, ...}$) and combines them using logical connectives such as `$\land$' (conjunction), `$\lor$' (disjunction), `$\to$' (implication), `$\leftrightarrow$' (double implication), and `$\neg$' (negation) to capture the relation between them. Propositional logic also defines a set of inference rules that can be used to draw a conclusion when given a set of propositions. For example, the inference rule Modus Ponens states that given the premise $((p \to q) \land p)$ read in natural language as "if p, then q, and p is true," we can conclude that q is true $((p \to q) \land p) \vdash q$. We explore seven different inference rules of propositional logic and their respective extensions in first-order logic as shown in table \ref{tab:classical_inference_rules}.

PL serves as a foundational framework for reasoning about truth values of statements, represented as propositions denoted by symbols like ${p, q, r, etc}$. Employing logical connectives such as `$\land$' (conjunction), `$\lor$' (disjunction), and `$\to$' (implication), it establishes relationships between these propositions. PL incorporates various inference rules, guiding the derivation of conclusions from given propositions. For instance, \textit{Modus Ponens} is an example of such inference rules where if presented with the premises $((p \to q) \land p)$—interpreted as ``if p, then q, and p is true''—we can deduce the truth of q, denoted as $((p \to q) \land p) \vdash q$. 

%In this work, we delve into eight distinct inference rules of propositional logic, detailed in Table \ref{tab:classical_inference_rules}.

\paragraph{First-order Logic (FOL)}
% First-order logic extends the simple propositional logic by using predicates instead of straightforward propositions. First-order logic introduces quantifiers (universal quantifier - for all ($\forall$) and existential quantifier - exists ($\exists$)), which allows the use of quantified variables to represent the non-logical objects. For example, we can write, "There exists x such that x is John and x is a student." to represent the simple propositional logic equivalent "John is a student". First-order logic also extends the inference rules of propositional logic in the same way; for instance, the Modus Ponens rule states that from the premise $\forall(p(x) \rightarrow q(x))$ and $p(a)$, we conclude $q(a)$ (e.g., ``All babies cry” and ``Jack is a baby”, we can conclude ``Jack cries”. We explore the same set of inference rules and combinations of inference rules for both propositional and first-order logic, as shown in table \ref{tab:classical_inference_rules}. For our data creation process, as shown further, we combine these inference rules to generate 25 diverse multi-step logical combinations of varying depths.

FOL builds upon the foundations of PL by introducing predicates and quantifiers. Predicates allow us to express relationships involving variables, and quantifiers such as the universal ($\forall$) and existential ($\exists$) quantifiers enable us to make statements about all or some elements in a domain. For instance, instead of stating ``John is a student,'' we can express it in FOL as ``There exists x such that x is John and x is a student.'' This logic extends the rules of PL, such as the \textit{Modus Ponens} rule, which lets us infer conclusions for specific instances from general premises.

%Table \ref{tab:classical_inference_rules} details the inference rules explored in our study. We leverage a set of these inference rules, applying them to both propositional and first-order logic to generate diverse multi-step logical combinations for our data creation process.
\section{Combinations of rules for Monotonic Logic} \label{app:Rule_Combinations}
\def\stackalignment{l}
\begin{table*}[htbp]
\centering
\resizebox{0.9\linewidth}{!}{
\begin{tabular}{llcc}
\toprule
\textbf{Rule Combinations} & \textbf{Premises in Story} & \textbf{Premise in Question} & \textbf{Answer} \\
\midrule \midrule

\begin{tabular}[c]{@{}l@{}}
\textbf{DS:} (P $\lor$ Q) $\land$ $\lnot$P  $\vdash$ Q \\
\textbf{MP: }(Q $\to$ R) $\land$ Q $\vdash$ R 
\end{tabular}
& 
(P $\lor$ Q), (Q $\to$ R) 
& 
$\lnot$P & 
\textbf{R: \checkmark}
% Is R true? \textbf{: Yes}
\\ \midrule

\begin{tabular}[c]{@{}l@{}}
\textbf{MT: }(P $\to$ Q) $\land$ $\lnot$Q  $\vdash$ $\lnot$P \\
\textbf{DS: } (P $\lor$ R) $\land$ $\lnot$P $\vdash$ R 
\end{tabular}
&
(P $\to$ Q), (P $\lor$ R)
& $\lnot$Q & 
\textbf{R: \checkmark}
% Is R true? \textbf{: Yes}
\\ \midrule

\begin{tabular}[c]{@{}l@{}}
\textbf{HS: }(P $\to$ Q) $\land$ (Q $\to$ R) $\vdash$ (P $\to$ R) \\
\textbf{MP: }(P $\to$ R) $\land$ P $\vdash$ R
\end{tabular}
& 
(P $\to$ Q), (Q $\to$ R) & 
P & 
\textbf{R: \checkmark}
% Is R true? \textbf{: Yes} 
\\ \midrule

\begin{tabular}[c]{@{}l@{}}
\textbf{CD: }(P $\to$ Q) $\land$ (R $\to$ S) $\land$ (P $\lor$ R) $\vdash$ (Q $\lor$ S) \\ 
\textbf{DS: }(Q $\lor$ S) $\land$ $\lnot$Q $\vdash$ S 
\end{tabular}
& 
\begin{tabular}[c]{@{}l@{}}
(P $\to$ Q), \\ 
(R $\to$ S), (P $\lor$ R) 
\end{tabular}
& 
$\lnot$Q & 
\textbf{S: \checkmark}
% Is S true? \textbf{: Yes}
\\ \midrule

\begin{tabular}[c]{@{}l@{}}
\textbf{DD: }(P $\to$ Q) $\land$ (R $\to$ S) $\land$ ($\lnot$Q $\lor$ $\lnot$S) $\vdash$ ($\lnot$P $\lor$ $\lnot$R) \\
 \textbf{DS: }($\lnot$P $\lor$ $\lnot$R) $\land$ P $\vdash$ $\lnot$R
\end{tabular}
& 
\begin{tabular}[c]{@{}l@{}}
(P $\to$ Q), \\
(R $\to$ S), ($\lnot$Q $\lor$ $\lnot$S) 
\end{tabular}
& 
P & 
\textbf{R: \xmark}
% Is R true? \textbf{: No}
\\ \midrule

\begin{tabular}[c]{@{}l@{}}
\textbf{BD: }(P $\to$ Q) $\land$ (R $\to$ S) $\land$ (P $\lor$ $\lnot$S) $\vdash$ (Q $\lor$ $\lnot$R) \\
\textbf{DS: }(Q $\lor$ $\lnot$R) $\land$ $\lnot$Q $\vdash$ $\lnot$R
\end{tabular}
& 
\begin{tabular}[c]{@{}l@{}}
(P $\to$ Q), \\
(R $\to$ S), (P $\lor$ $\lnot$S) 
\end{tabular}
& 
$\lnot$Q & 
\textbf{R: \xmark}
% Is R true? \textbf{: No}
\\ \midrule

\begin{tabular}[c]{@{}l@{}}
\textbf{HS: }(P $\to$ Q) $\land$ (Q $\to$ R) $\vdash$ (P $\to$ R) \\
\textbf{MT: }(P $\to$ R) $\land$ $\lnot$R $\vdash$ $\lnot$P
\end{tabular}
&
(P $\to$ Q), (Q $\to$ R) & 
$\lnot$R & 
\textbf{P: \xmark}
% Is P true? \textbf{: No}
\\ 
\bottomrule
\end{tabular}
}
\caption{2-step reasoning rule combinations for PL and FOL.}
\label{tab:two_step_reasoning_combinations}
\end{table*}
We created 27 multi-step reasoning inference rule combinations for Propositional Logic (PL), with depths ranging from 2 to 5. We use the same rule combinations for First Order Logic (FOL) for each depth. All rule combinations for 2-step, 3-step, 4-step, and 5-step reasoning for PL and FOL are presented in Tables \ref{tab:two_step_reasoning_combinations}, \ref{tab:three_step_reasoning_combinations}, \ref{tab:four_step_reasoning_combinations}, and \ref{tab:five_step_reasoning_combinations} respectively. For each combination, we provide the inference rules to be used for reasoning, the premises present in the context and in the question, and the complex reasoning question-answer pair.

\def\stackalignment{l}
\begin{table*}[htbp]
\centering
\resizebox{0.9\linewidth}{!}{
\begin{tabular}{llcc}
\toprule
\textbf{Rule Combinations} & \textbf{Premises in Story} & \textbf{Premise in Question} & \textbf{Answer} \\
\midrule \midrule

\begin{tabular}[c]{@{}l@{}}
\textbf{HS: }(P $\to$ Q) $\land$ (Q $\to$ R) $\vdash$ (P $\to$ R) \\
\textbf{MP: }(P $\to$ R) $\land$ P $\vdash$ R \\ 
\textbf{MP: }(R $\to$ S) $\land$ R $\vdash$ S
\end{tabular}
& 
\begin{tabular}[c]{@{}l@{}}
(P $\to$ Q), \\
(Q $\to$ R), (R $\to$ S) 
\end{tabular}
& 
P & 
\textbf{S: \checkmark}
% Is S true? \textbf{: Yes} 
\\ \midrule

\begin{tabular}[c]{@{}l@{}}
\textbf{CD: }(P $\to$ Q) $\land$ (R $\to$ S) $\land$ (P $\lor$ R) $\vdash$ (Q $\lor$ S) \\
\textbf{DS: }(Q $\lor$ S) $\land$ $\lnot$Q $\vdash$ S\\
\textbf{MP: }(S $\to$ T) $\land$ S $\vdash$ T
\end{tabular}
& 
\begin{tabular}[c]{@{}l@{}}
(P $\to$ Q), (R $\to$ S), \\
(P $\lor$ R), (S $\to$ T) 
\end{tabular}
& 
$\lnot$Q & 
\textbf{T: \checkmark}
% Is T true? \textbf{: Yes}
\\ \midrule

\begin{tabular}[c]{@{}l@{}}
\textbf{BD: }(P $\to$ Q) $\land$ (R $\to$ S) $\land$ (P $\lor$ $\lnot$S) $\vdash$ (Q $\lor$ $\lnot$R) \\ 
\textbf{CT: }(Q $\lor$ $\lnot$R) $\dashv$$\vdash$ ($\lnot$R $\lor$ Q) \\ \textbf{DS: }($\lnot$R $\lor$ Q) $\land$ R $\vdash$ Q
\end{tabular}
& 
\begin{tabular}[c]{@{}l@{}}
(P $\to$ Q), \\
(R $\to$ S), (P $\lor$ $\lnot$S)
\end{tabular}
& 
R & 
\textbf{Q: \checkmark}
% Is Q true? \textbf{: Yes}
\\ \midrule

\begin{tabular}[c]{@{}l@{}}
\textbf{BD: }(P $\to$ Q) $\land$ (R $\to$ S) $\land$ (P $\lor$ $\lnot$S) $\vdash$ (Q $\lor$ $\lnot$R) \\ \textbf{DS: }(Q $\lor$ $\lnot$R) $\land$ $\lnot$Q $\vdash$ $\lnot$R \\ \textbf{MT: }(T $\to$ R) $\land$ $\lnot$R  $\vdash$ $\lnot$T
\end{tabular}
& 
\begin{tabular}[c]{@{}l@{}}
(P $\to$ Q), (R $\to$ S), \\
(P $\lor$ $\lnot$S), (T $\to$ R) 
\end{tabular}
& 
$\lnot$Q & 
\textbf{T: \xmark}
% Is T true? \textbf{: No}
\\ \midrule

\begin{tabular}[c]{@{}l@{}}
\textbf{CD: }(P $\to$ Q) $\land$ (R $\to$ S) $\land$ (P $\lor$ R) $\vdash$ (Q $\lor$ S) \\ \textbf{CT: }(Q $\lor$ S) $\dashv$$\vdash$ (S $\lor$ Q) \\ \textbf{DS: }(S $\lor$ Q) $\land$ $\lnot$S $\vdash$ Q
\end{tabular}
& 
\begin{tabular}[c]{@{}l@{}}
(P $\to$ Q), \\
(R $\to$ S), (P $\lor$ R)
\end{tabular}& 
$\lnot$S & 
\textbf{Q: \checkmark}
% Is Q true? \textbf{: Yes}
\\ \midrule

\begin{tabular}[c]{@{}l@{}}
\textbf{HS: }(P $\to$ Q) $\land$ (Q $\to$ R) $\vdash$ (P $\to$ R) \\ \textbf{CD: }(P $\to$ R) $\land$ (S $\to$ T) $\land$ (P $\lor$ S) $\vdash$ (R $\lor$ T) \\ \textbf{DS: }(R $\lor$ T) $\land$ $\lnot$R $\vdash$ T
\end{tabular}
& 
\begin{tabular}[c]{@{}l@{}}
(P $\to$ Q), (Q $\to$ R), \\
(S $\to$ T), (P $\lor$ S) 
\end{tabular}
& 
$\lnot$R & 
\textbf{T: \checkmark}
% Is T true? \textbf{: Yes}
\\ \midrule

\begin{tabular}[c]{@{}l@{}} 
\textbf{HS: }(P $\to$ Q) $\land$ (Q $\to$ R) $\vdash$ (P $\to$ R) \\ \textbf{MT: }(P $\to$ R) $\land$ $\lnot$R $\vdash$ $\lnot$P \\ \textbf{DS: }(P $\lor$ S) $\land$ $\lnot$P $\vdash$ S 
\end{tabular}
&
\begin{tabular}[c]{@{}l@{}}
(P $\to$ Q), \\
(Q $\to$ R), (P $\lor$ S) 
\end{tabular} & 
$\lnot$R & 
\textbf{S: \checkmark}
% Is S true? \textbf{: Yes}
\\ \midrule

\begin{tabular}[c]{@{}l@{}}
\textbf{DD: }(P $\to$ Q) $\land$ (R $\to$ S) $\land$ ($\lnot$Q $\lor$ $\lnot$S) $\vdash$ ($\lnot$P $\lor$ $\lnot$R) \\ 
\textbf{DS: }($\lnot$P $\lor$ $\lnot$R) $\land$ P $\vdash$ $\lnot$R \\ \textbf{MT: }(T $\to$ R) $\land$ $\lnot$R $\vdash$ $\lnot$T 
\end{tabular}
& 
\begin{tabular}[c]{@{}l@{}}
(P $\to$ Q), (R $\to$ S),  \\
($\lnot$Q $\lor$ $\lnot$S), (T $\to$ R) 
\end{tabular}
& 
P & 
\textbf{T: \xmark}
% Is T true? \textbf{: No}
\\ \midrule

\begin{tabular}[c]{@{}l@{}}
\textbf{DMT: }($\lnot$Q $\lor$ $\lnot$R) $\dashv\vdash$ $\lnot$(Q $\land$ R)   \\ 
\textbf{CO: }(P $\to$ Q) $\land$ (P $\to$ R) $\vdash$  P $\to$ (Q $\land$ R) \\ \textbf{MT: }(P $\to$ (Q $\land$ R) $\land$ $\lnot$(Q $\land$ R) $\vdash$ $\lnot$P 
\end{tabular}
& 
\begin{tabular}[c]{@{}l@{}}
(P $\to$ Q), (P $\to$ R)
\end{tabular}
& 
$\neg$Q $\lor$ $\neg$R& 
\textbf{P: \xmark}
% Is T true? \textbf{: No}
\\ 

\bottomrule
\end{tabular}
}
\caption{3-step reasoning rule combinations for PL and FOL.}
\label{tab:three_step_reasoning_combinations}
\end{table*}
\def\stackalignment{l}

\begin{table*}[htbp]
\centering
\resizebox{0.9\linewidth}{!}{
\begin{tabular}{llcc}
\toprule
\textbf{Rule Combinations} & \textbf{Premises in Story} & \textbf{Premise in Question} & \textbf{Answer} \\
\midrule \midrule

\begin{tabular}[c]{@{}l@{}}
\textbf{CD: }(P $\to$ Q) $\land$ (R $\to$ S) $\land$ (P $\lor$ R) $\vdash$ (Q $\lor$ S) \\
\textbf{DS: }(Q $\lor$ S) $\land$ $\lnot$Q $\vdash$ S \\ 
\textbf{MP: }(S $\to$ T) $\land$ S $\vdash$ T \\ 
\textbf{MP: }(T $\to$ U) $\land$ T $\vdash$ U
\end{tabular}
& 
\begin{tabular}[c]{@{}l@{}}
(P $\to$ Q), \\
(R $\to$ S), (P $\lor$ R), \\
(S $\to$ T), (T $\to$ U) 
\end{tabular}
& 
$\lnot$Q & 
\textbf{U: \checkmark}
% Is U true? \textbf{: Yes}
\\ \midrule

\begin{tabular}[c]{@{}l@{}}
\textbf{BD: }(P $\to$ Q) $\land$ (R $\to$ S) $\land$ (P $\lor$ $\lnot$S) $\vdash$ (Q $\lor$ $\lnot$R) \\
\textbf{CT: }(Q $\lor$ $\lnot$R) $\dashv$$\vdash$ ($\lnot$R $\lor$ Q) \\
\textbf{DS: }($\lnot$R $\lor$ Q) $\land$ R $\vdash$ Q \\ 
\textbf{MP: }(Q $\to$ T) $\land$ Q $\vdash$ T
\end{tabular}
& 
\begin{tabular}[c]{@{}l@{}}
(P $\to$ Q), (R $\to$ S), \\
(P $\lor$ $\lnot$S), (Q $\to$ T) 
\end{tabular}
& 
R & 
\textbf{T: \checkmark}
% Is T true? \textbf{: Yes}
\\ \midrule

\begin{tabular}[c]{@{}l@{}}
\textbf{BD: }(P $\to$ Q) $\land$ (R $\to$ S) $\land$ (P $\lor$ $\lnot$S) $\vdash$ (Q $\lor$ $\lnot$R) \\ 
\textbf{DS: }(Q $\lor$ $\lnot$R) $\land$ $\lnot$Q $\vdash$ $\lnot$R \\ 
\textbf{MT: }(T $\to$ R) $\land$ $\lnot$R  $\vdash$ $\lnot$T \\ \textbf{DS: }(T $\lor$ U) $\land$ $\lnot$T $\vdash$ U
\end{tabular} 
& 
\begin{tabular}[c]{@{}l@{}}
(P $\to$ Q), \\
(R $\to$ S), (P $\lor$ $\lnot$S), \\
(T $\to$ R), (T $\lor$ U) 
\end{tabular}
& 
$\lnot$Q & 
\textbf{U: \checkmark}
% Is U true? \textbf{: Yes}
\\ \midrule

\begin{tabular}[c]{@{}l@{}}
\textbf{HS: }(P $\to$ Q) $\land$ (Q $\to$ R) $\vdash$ (P $\to$ R) \\ \textbf{CD: }(P $\to$ R) $\land$ (S $\to$ T) $\land$ (P $\lor$ S) $\vdash$ (R $\lor$ T) \\ \textbf{DS: }(R $\lor$ T) $\land$ $\lnot$R $\vdash$ T \\ \textbf{MP: }(T $\to$ U) $\land$ T $\vdash$ U 
\end{tabular} 
& 
\begin{tabular}[c]{@{}l@{}}
(P $\to$ Q), \\
(Q $\to$ R), (S $\to$ T), \\
(P $\lor$ S), (T $\to$ U) 
\end{tabular}
& 
$\lnot$R & 
\textbf{U: \checkmark}
% Is U true? \textbf{: Yes}
\\ \midrule

\begin{tabular}[c]{@{}l@{}}
\textbf{CD: }(P $\to$ Q) $\land$ (R $\to$ S) $\land$ (P $\lor$ R) $\vdash$ (Q $\lor$ S) \\ 
\textbf{CT: }(Q $\lor$ S) $\dashv$$\vdash$ (S $\lor$ Q) \\ \textbf{DS: }(S $\lor$ Q) $\land$ $\lnot$S $\vdash$ Q \\ 
\textbf{MP: }(Q $\to$ T) $\land$ Q $\vdash$ T
\end{tabular}
& 
\begin{tabular}[c]{@{}l@{}}
(P $\to$ Q), (R $\to$ S), \\
(P $\lor$ R), (Q $\to$ T) 
\end{tabular}
& 
$\lnot$S & 
\textbf{T: \checkmark}
% Is T true? \textbf{: Yes}
\\ \midrule

\begin{tabular}[c]{@{}l@{}}
\textbf{HS: }(P $\to$ Q) $\land$ (Q $\to$ R) $\vdash$ (P $\to$ R) \\ \textbf{MT: }(P $\to$ R) $\land$ $\lnot$R $\vdash$ $\lnot$P\\ 
\textbf{DS: }(P $\lor$ S) $\land$ $\lnot$P $\vdash$ S \\ 
\textbf{MP: }(S $\to$ T) $\land$ S $\vdash$ T
\end{tabular}
&
\begin{tabular}[c]{@{}l@{}}
(P $\to$ Q), (Q $\to$ R), \\
(P $\lor$ S), (S $\to$ T) 
\end{tabular} 
& 
$\lnot$R & 
\textbf{T: \checkmark}
% Is T true? \textbf{: Yes}
\\ \midrule

\begin{tabular}[c]{@{}l@{}}
\textbf{BD: }(P $\to$ Q) $\land$ (R $\to$ S) $\land$ (P $\lor$ $\lnot$S) $\vdash$ (Q $\lor$ $\lnot$R) \\ 
\textbf{DS: }(Q $\lor$ $\lnot$R) $\land$ $\lnot$Q $\vdash$ $\lnot$R \\
\textbf{MT: }(T $\to$ R) $\land$ $\lnot$R  $\vdash$ $\lnot$T \\ \textbf{MT: }(U $\to$ T) $\land$ $\lnot$T  $\vdash$ $\lnot$U
\end{tabular}
&
\begin{tabular}[c]{@{}l@{}}
(P $\to$ Q), \\
(R $\to$ S), (P $\lor$ $\lnot$S), \\
(T $\to$ R), (U $\to$ T) 
\end{tabular}
& 
$\lnot$Q & 
\textbf{U: \xmark}
% Is U true? \textbf{: No}
\\ \midrule

\begin{tabular}[c]{@{}l@{}}
\textbf{IM: }(P $\to$ (Q $\land$ R)) $\vdash$ (P $\land$ Q) $\to$ R \\ 
\textbf{MT: }((P $\land$ Q) $\to$ R) $\land$ $\lnot$R $\vdash$ $\lnot$(P $\land$ Q) \\
\textbf{DMT: }$\lnot$(P $\land$ Q) $\vdash$ ($\lnot$P $\lor$ $\lnot$Q) \\ \textbf{DS: }($\lnot$P $\lor$ $\lnot$Q) $\land$Q  $\vdash$ $\lnot$P
\end{tabular}
&
\begin{tabular}[c]{@{}l@{}}
(P $\to$ (Q $\land$ R))
\end{tabular}
& 
Q, $\lnot$R & 
\textbf{P: \xmark}
% Is U true? \textbf{: No}
\\ 

\bottomrule
\end{tabular}
}
\caption{4-step reasoning rule combinations for PL and FOL.}
\label{tab:four_step_reasoning_combinations}
\end{table*}
\def\stackalignment{l}

\begin{table*}[htbp]
\centering
\resizebox{0.9\linewidth}{!}{
\begin{tabular}{llcc}
\toprule
\textbf{Rule Combinations} & \textbf{Premises in Story} & \textbf{Premise in Question} & \textbf{Answer} \\
\midrule \midrule 

\begin{tabular}[c]{@{}l@{}}
\textbf{HS: }(P $\to$ Q) $\land$ (Q $\to$ R) $\vdash$ (P $\to$ R) \\
\textbf{MT: }(P $\to$ R) $\land$ $\lnot$R $\vdash$ $\lnot$P \\ 
\textbf{DS: }(P $\lor$ S) $\land$ $\lnot$P $\vdash$ S \\ 
\textbf{MP: }(S $\to$ T) $\land$ S $\vdash$ T \\
\textbf{MP: }(T $\to$ U) $\land$ T $\vdash$ U
\end{tabular} 
&
\begin{tabular}[c]{@{}l@{}}
(P $\to$ Q), \\
(Q $\to$ R), (P $\lor$ S), \\
(S $\to$ T), (T $\to$ U) 
\end{tabular}
& 
$\lnot$R & 
\textbf{U: \checkmark}
% Is U true? \textbf{: Yes}
\\ \midrule

\begin{tabular}[c]{@{}l@{}}
\textbf{BD: }(P $\to$ Q) $\land$ (R $\to$ S) $\land$ (P $\lor$ $\lnot$S) $\vdash$ (Q $\lor$ $\lnot$R) \\ \textbf{CT: }(Q $\lor$ $\lnot$R) $\dashv$$\vdash$ ($\lnot$R $\lor$ Q) \\ \textbf{DS: }($\lnot$R $\lor$ Q) $\land$ R $\vdash$ Q \\ \textbf{MP: }(Q $\to$ T) $\land$ Q $\vdash$ T \\ \textbf{MP: }(T $\to$ U) $\land$ T $\vdash$ U
\end{tabular}
&
\begin{tabular}[c]{@{}l@{}}
(P $\to$ Q), \\
(R $\to$ S), (P $\lor$ $\lnot$S), \\
(Q $\to$ T), (T $\to$ U) 
\end{tabular}
& 
R & 
\textbf{U: \checkmark}
% Is U true? \textbf{: Yes}
\\ \midrule

\begin{tabular}[c]{@{}l@{}}
\textbf{CD: }(P $\to$ Q) $\land$ (R $\to$ S) $\land$ (P $\lor$ R) $\vdash$ (Q $\lor$ S) \\ 
\textbf{CT: }(Q $\lor$ S) $\dashv$$\vdash$ (S $\lor$ Q) \\ 
\textbf{DS: }(S $\lor$ Q) $\land$ $\lnot$S $\vdash$ Q \\ 
\textbf{MP: }(Q $\to$ T) $\land$ Q $\vdash$ T \\ 
\textbf{MP: }(T $\to$ U) $\land$ T $\vdash$ U
\end{tabular}
&
\begin{tabular}[c]{@{}l@{}}
(P $\to$ Q), \\ 
(R $\to$ S), (P $\lor$ R), \\
(Q $\to$ T), (T $\to$ U) 
\end{tabular}& 
$\lnot$S & 
\textbf{U: \checkmark}
% Is U true? \textbf{: Yes}
\\ 

\bottomrule
\end{tabular}
}
\caption{5-step reasoning rule combinations for PL and FOL.}
\label{tab:five_step_reasoning_combinations}
\end{table*}

\section{Example of Prompt}\label{app:prompt}
\begin{figure}[ht]
    \centering
    \includegraphics[width=0.96\linewidth]{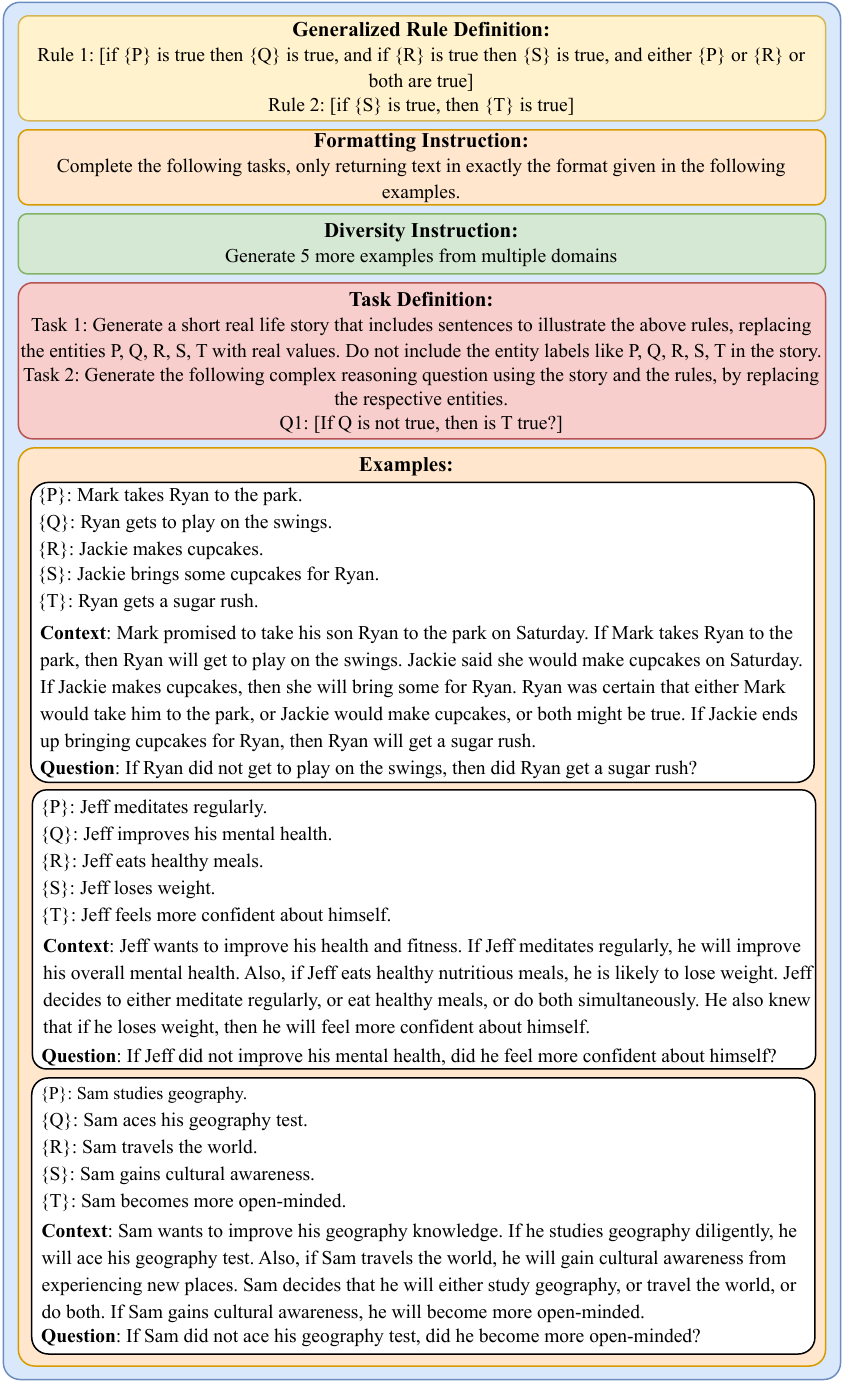}
    \caption{An example prompt for 3-step combination of inference rules CD, DS, and MP from propositional logic.}
    \label{fig:extend_prompt}
\end{figure}

Figure \ref{fig:extend_prompt} illustrates an example prompt for combination of rules from propositional logic, namely `constructive dilemma' (CD), `disjunctive syllogism' (DS), and `modus ponens' (MP). CD is represented as $(p\to q)\land (r\to s)\land (p\lor r))\vdash (q\lor s)$, which can be understood in natural language as ``If $p$ implies $q$, and if $r$ implies $s$, and either $p$ or $r$ or both are true, then we can conclude that either $q$ or $s$ or both are true.'' DS is formally represented as $(p\lor q)\land \neg p)\vdash q$, which can be understood in natural language as ``If $p$ or $q$ are true, and we know $\neg p$, then we can conclude $q$.'' MP is formally represented as $(p \to q) \land p) \vdash q $, which can be understood in natural language as ``If $p$ implies $q$, and we know $p$, then we can conclude $q$.'' 

In this prompt, the generalized rule definitions provide a description of the premises given in the story in natural language. The prompt includes instructions on how the generated samples should be formatted, instructions to generate samples from diverse domains, and detailed definitions for generating propositions, and then using them to generate a context and question for each sample. To enhance the quality of samples in terms of relevance and coherence, the prompt includes an examples section that demonstrates these tasks. In Figure \ref{fig:extend_prompt}, we present three examples with their respective propositions, contexts, and questions. 

\section{NL Examples for PL and FOL}\label{app:NL_examples}
\providecommand{\Mohith}[1]{
    {\protect\color{blue}{[Mohith: #1]}}
}

\begin{table*}[ht]
\centering
\resizebox{0.9\linewidth}{!}{
\begin{tabular}{c|m{6cm}|m{12cm}}
\toprule
\multicolumn{1}{c|}{Depth} & \multicolumn{1}{c|}{Rules and Propositions}    & \multicolumn{1}{c}{Context and Question} \\ \midrule \midrule

2 &
\textbf{Rules:} CD, DS  \newline \newline \textbf{Propositions:} \newline P: There is a big snowstorm coming.\newline Q: Schools will be closed.\newline R: Boss tells us to work from home.\newline S: I avoid driving in the snow & 
\textbf{Context:} If there is a big snowstorm coming, schools will be closed tomorrow. Also, if my boss tells us to work from home, I can avoid driving in the snow. It seems either there will be a snowstorm or I'll be told to work from home, maybe both. \newline \newline \textbf{Question:} If schools were not closed tomorrow, then did I avoid driving in the snow?\\ \midrule

3 &
\textbf{Rules:} BD, DS, MT  \newline \newline 
\textbf{Propositions:} \newline P: The weather is nice.\newline Q: She goes for a walk.\newline R: Finishes chores.\newline S: Has free time.\newline T: It's the weekend & 
\textbf{Context:} It was a beautiful sunny day. Amy knew that if the weather is nice, she goes for a walk. Amy also had chores to complete today. If Amy finishes her chores, then she has free time. Amy is certain that either the weather is nice, or she doesn't have free time, or the weather is nice and she doesn't have free time. She also knows that if it's the weekend, then she finishes her chores. \newline \newline \textbf{Question:} If Amy didn't go for a walk, then is it the weekend?\\ \midrule

4 &
\textbf{Rules:} HS, CD, DS, MP  \newline \newline 
\textbf{Propositions:} \newline P: Studied hard for the exam.\newline Q: Feel confident.\newline R:  Score well.\newline S: Cooked nice dinner.\newline T: Feel relaxed.\newline U: Sleep soundly. & 
\textbf{Context:} Jim had a big exam coming up that he needed to prepare for. If Jim studied hard for the exam, he would feel confident going into it. If Jim felt confident about the exam, he would end up scoring well on it. His wife Lucy enjoyed cooking nice dinners. If Lucy cooked a nice dinner, she felt relaxed afterwards. Last night, either Jim studied hard, or Lucy cooked a nice dinner, or they both did those things. Jim knew that if Lucy felt relaxed after dinner, she always slept soundly through the night. \newline \newline \textbf{Question:} If Jim did not score well on the exam, did Lucy sleep soundly?\\ \midrule

5 &
\textbf{Rules:} HS, MT, DS, MP, MP  \newline \newline
\textbf{Propositions:} \newline P: Train consistently. \newline Q: Increase endurance and stamina. \newline R: complete the 26.2 mile marathon.\newline S: Ate nutritious food.\newline T: More steady energy.\newline U: Train harder staying injury free. & 
\textbf{Context:} Jessica set a goal to run a marathon. She learned that if she trained consistently, she could increase her endurance and stamina. Jessica knew that if her endurance improved, she could complete the 26.2 mile marathon. To complement her training, Jessica made sure she either trained regularly, or ate nutritious foods, or did both. Eating nutritious foods gave Jessica more steady energy for her workouts. With this extra energy, Jessica found she could train harder while staying injury-free on her road to marathon success.\newline \newline \textbf{Question:} If Jessica does not complete the marathon, then does she stay injury-free during training?\\ 

\bottomrule

\end{tabular}
}
\caption{Natural language examples of rule combinations of each depth for PL.}
\label{tab:NL_examples_PL}
\end{table*}

\begin{table*}[ht]
\centering
\resizebox{0.9\linewidth}{!}{
\begin{tabular}{c|m{7cm}|m{13cm}}
\toprule
\multicolumn{1}{c|}{Depth} & \multicolumn{1}{c|}{Rules and Predicates}    & \multicolumn{1}{c}{Context and Question} \\ \midrule \midrule

2 &
\textbf{Rules:} CD, DS  \newline \newline  
\textbf{Predicates:} \newline P: Compose original music.\newline Q: Work would be Unique. \newline R: Promote music online. \newline S: Gain following. & 
\textbf{Context:} An aspiring musician decided to try writing their own songs. They realized that if they composed original music, their work would be unique; if they promoted their music online, they would gain a following. The musician could write original songs or promote their music online. \newline \newline \textbf{Question:} Given that Maria's music was not unique, is it true that she gained a following online?\\ \midrule

3 &
\textbf{Rules:} BD, DS, MT  \newline \newline 
\textbf{Predicates:} \newline P: It's Monday. \newline Q: There is a staff meeting. \newline R: Finish report. \newline S: Submit the report.\newline T: Good Employee. & 
\textbf{Context:} It was a busy morning at the office. If it was Monday, then there would be a staff meeting. If they finished the report, then they could submit it to their manager. They were certain that either it was Monday, or they did not submit the report. It is known at the office that if someone is a good employee, they finish their reports on time.\newline \newline \textbf{Question:} Sam did not have a staff meeting, is Sam a good employee?\\ \midrule

4 &
\textbf{Rules:} BD, DS, MT, DS  \newline \newline \textbf{Predicates:} \newline P: First day of school.\newline Q: students feel nervous and excited.\newline R:  Study Hard.\newline S: get good grades.\newline T: teacher is very strict.\newline U: class textbook is very long. & 
\textbf{Context:} If it is the first day of school, then students feel nervous and excited. If someone studies hard, then they get good grades. Either it is the first day, or they do not get good grades, or it is the first day and they do not get good grades. If a teacher is very strict, then students have to study hard for that class. Either the teacher is very strict, or the class textbook is very long, or perhaps both are true. \newline \newline \textbf{Question:} Emma was not nervous on the first day, does this mean did she have a very long textbook in one of her classes?\\ \midrule

5 &
\textbf{Rules:} HS, MT, DS, MP, MP  \newline \newline \textbf{Predicates:} \newline P: Practice drawing techniques.\newline Q: improve artistic skills. \newline R: sell their artwork.\newline S: studies art history and famous artists.\newline T: gain inspiration.\newline U: develop creative style. & 
\textbf{Context:} Someone wanted to become an artist. They learned that if they practiced drawing techniques consistently, they would improve their artistic skills. With improved artistic skills, they could sell their artworks. Either someone practices drawing techniques consistently, or someone studies art history and famous artists, or they do both. If someone studies art history and famous artists, then they gain inspiration for their own art. If they gain inspiration, then they can develop their own creative style.\newline \newline \textbf{Question:} If Emma cannot sell her artworks yet, then has she developed her own creative style?\\ 

\bottomrule

\end{tabular}
}
\caption{Natural language examples of rule combinations of each depth for FOL.}
\label{tab:NL_examples_FOL}
\end{table*}

In this section, we illustrate multi-step reasoning for PL and FOL using natural language examples for depths 2 through 5. Table \ref{tab:NL_examples_PL} provides examples in natural language for PL. We provide one example of rule combinations for each depth. For each example, we provide the inference rules and propositions, as well as the respective context and complex reasoning question. Table \ref{tab:NL_examples_FOL} provides examples in natural language for FOL, with one combination for each depth. Similar to PL, we provide the inference rules, predicates, and the context-question pair for each example.

\section{More Details on NM} \label{app:NM}
\begin{table*}[htbp]
  \centering
  \resizebox{0.9\linewidth}{!}{
    \begin{tabular}{c|c} \toprule
    \textbf{Rule} & \textbf{Extended First-order Logic with Multi-variable}  \\ \midrule \midrule
    \begin{tabular}[c]{@{}c@{}}1\end{tabular}  & $\forall x\forall y((p(x) \land q(x)) \to r(x,y)) \land \exists u \exists v(p(u) \land \neg r(u,v)) \vdash \exists y \neg q(y) $\\ \midrule
    
    \begin{tabular}[c]{@{}c@{}}2\end{tabular} & $\forall x\forall y((p(x) \land q(x)) \to \neg s(x,y)) \land \forall z(r(z) \to p(z)) \land r(a) \land s(a,b) \vdash \neg q(b) $  \\ \midrule
    
    \begin{tabular}[c]{@{}c@{}}3\end{tabular} & $\forall x\exists y((p(x) \to q(x,y)) \land \forall u \forall v ((q(u,v) \land r(u,v)) \to s(v)) \land \exists z \exists k(p(z) \land r(z,k)) \vdash \exists w s(w) $ \\ \midrule
    
    \begin{tabular}[c]{@{}c@{}}4\end{tabular} & $\forall x\forall y\forall z(p(x,y,z) \to (q(x,z)\lor r(y))) \land \exists u \exists v \exists w(p(u,v,w) \land \neg q(u,w)) \vdash \exists sr(s) $ \\ \midrule
    
    \begin{tabular}[c]{@{}c@{}}5\end{tabular} & $\forall x((p(x) \to \exists yr(y,x)) \land p(a) \vdash \exists z r(z,a) $\\ \midrule
    
    \begin{tabular}[c]{@{}c@{}}6\end{tabular} & $\forall x\forall y(p(x,y) \lor q(x,y)) \land  \exists u\exists v \neg q(u,v) \vdash \exists z\exists wp(z,w) $ \\ \midrule
    
    \begin{tabular}[c]{@{}c@{}}7\end{tabular} & $\forall x\forall y(p(x,y) \to (q(x) \land r(y)) \land p(a,b) \vdash q(a) \land r(b)) $ \\ \bottomrule
    
    \end{tabular}%
}
    \caption{FOL inference rules that establish the relationship between multiple variables}
    \label{tab:multivariable_inference_rules}%
\end{table*}

\begin{table*}[htbp]
\centering
\resizebox{0.9\linewidth}{!}{
\begin{tabular}{l|l}
\toprule
\multicolumn{1}{c|}{\textbf{Basic Default Reasoning}}  & \multicolumn{1}{c}{\textbf{Default Reasoning with Irrelevant Information}}      \\ \midrule

\begin{tabular}[c]{@{}l@{}}Context: Blocks A and B are heavy.\\                Heavy blocks are typically located on the table.\\                A is not on the table.\\  \\ Conclusion: B is on the table.\end{tabular}                   
& 
\begin{tabular}[c]{@{}l@{}}Context: Blocks A and B are heavy.\\                Heavy blocks are typically located on the table.\\                A is not on the table.\\                B is red.\\ \\  Conclusion: B is on the table.\end{tabular}       \\ \midrule

% Default Reasoning with Several Defaults (DRS), Default Reasoning with Irrelevant Information (DRI), Default Reasoning
% with a Disabled Default (DRD), Default Reasoning in an Open Domain (DRO), Reasoning about Unknown Expectations I (RE1), Reasoning about Unknown Expectations II (RE2), Reasoning about Unknown Expectations III (RE3), Reasoning about Priorities (RAP)

\multicolumn{1}{c|}{\textbf{Default Reasoning with a Disabled Default}}  & \multicolumn{1}{c}{\textbf{Default Reasoning in an Open Domain}}      \\ \midrule \midrule

\begin{tabular}[c]{@{}l@{}}Context: Block A and B are heavy \\                Heavy blocks are normally located on the table.\\                A is possibly an exception to this rule.\\  \\ Conclusion: B is on the table.\end{tabular}   
& 
\begin{tabular}[c]{@{}l@{}}Context: Block A is heavy.\\
Heavy blocks are normally located on the table.\\            A is not on the table. \\ \\ Conclusion: All heavy blocks other than A are on the table.\end{tabular}       \\ \midrule 

\multicolumn{1}{c|}{\textbf{Reasoning about Unknown Expectations I}}  & \multicolumn{1}{c}{\textbf{Reasoning about Unknown Expectations II}}      \\ \midrule \midrule

\begin{tabular}[c]{@{}l@{}}Context: Blocks A, B, and C are heavy. \\           Heavy blocks are normally located on the table.\\            
At least one of A, B, is not on the table.\\  \\ Conclusion:  C is on the table.\\ Exactly one of A, B is not on the table.
\end{tabular}                   
& 
\begin{tabular}[c]{@{}l@{}}Context: Heavy blocks are normally located on the table. \\ At least one heavy block is not on the table.\\ \\  Conclusion: Exactly one heavy block is not on the table. \end{tabular}       \\ 
\midrule

\multicolumn{1}{c|}{\textbf{Reasoning about Unknown Expectations III}} & \multicolumn{1}{c}{\textbf{Reasoning about Priorities}}     \\ \midrule \midrule

\begin{tabular}[c]{@{}l@{}}Context: Blocks A is heavy. \\
Heavy blocks are normally located on the table. \\
At least one heavy block is not on the table. \\ \\ Conclusion: A is on the table.\end{tabular} 
& 
\begin{tabular}[c]{@{}l@{}}Context: Jack asserts that block A is on the table.\\                Mary asserts that block A is not on the table.\\                When people assert something, they are normally right.\\ \\ Conclusion: If Mary’s evidence is more reliable than Jack’s.\\                     then block A is not on the table\end{tabular} \\ 

\bottomrule

\end{tabular}}
\caption{Illustrative examples of non-monotonic reasoning adapted from \cite{lifschitz1989benchmark}.}
\label{tab:nm_examples}
\end{table*}
% \usepackage[normalem]{ulem}
\begin{table*}
\centering
\resizebox{0.9\linewidth}{!}{
\begin{tabular}{m{4.5cm}|m{14cm}}
\toprule
Rule  & Examples \\ \midrule \midrule

\textbf{BDR\_MP} \newline \newline Conclusion of BDR: X \newline MP: (X $\to$ Y) $\land$ X $\vdash$ Y %(X → Y) ∧ X ⊢ Y
% Jim and Pam work at the same office. Normally, employees at that office get free lunch. Jim does not get free lunch. If Pam gets free lunch, then she gets an hour lunch break.",
%             "question": "Can we conclude Pam gets an hour lunch break?",

& \textbf{Context:} Jim and Pam work at the same office. Normally, employees at that office get free lunch. Jim does not get free lunch. If Pam gets free lunch, then she gets an hour lunch break. \newline \newline \textbf{Question:} Can we conclude Pam gets an hour lunch break? (Yes) \\ \midrule

\textbf{BDR\_MT} \newline \newline Conclusion of BDR: X \newline
% MP: (X $\to$ Y) $\land$ X $\vdash$ Y %(X → Y) ∧ X ⊢ Y
MT: (X $\to$ Y) $\land$ $\lnot$Y  $\vdash$ $\lnot$X % (X → Y) ∧ ¬Y ⊢ ¬X
& \textbf{Context:} Emma and Jacob are students in the same class. Usually students in that class submit homework assignments. Emma did not submit the last homework. If Jacob missed over 3 classes, that means he likely did not submit the homework. \newline\newline \textbf{Question:} Can we conclude Jacob missed over 3 classes? (No) \\ \midrule

\textbf{DRD\_MP} \newline\newline Conclusion of DRD: X \newline
% MT: (X $\to$ Y) $\land$ $\lnot$Y  $\vdash$ $\lnot$X % (X → Y) ∧ ¬Y ⊢ ¬X
MP: (X $\to$ Y) $\land$ X $\vdash$ Y %(X → Y) ∧ X ⊢ Y
& \textbf{Context:} The Honda and Toyota are sedans. Sedans normally have four doors. The Honda might not have four doors even though it's a sedan. If the Toyota has a four doors then it has four windows. \newline\newline \textbf{Question:} Can we conclude the Toyota likely has four windows? (Yes)   \\ \midrule

\textbf{DRD\_MT} \newline\newline Conclusion of DRD: X\newline
% MP: (X $\to$ Y) $\land$ X $\vdash$ Y %(X → Y) ∧ X ⊢ Y
MT: (X $\to$ Y) $\land$ $\lnot$Y  $\vdash$ $\lnot$X % (X → Y) ∧ ¬Y ⊢ ¬X
& \textbf{Context:} Oaks and pines are types of trees. Typically trees grow from seeds. Oaks may not grow from seeds even though they are trees. If a pine is artificial, then it does not grow from a seed.\newline\newline \textbf{Question:} Can we conclude the pine is artificial? (No)    \\ \midrule

\textbf{DRI\_MP} \newline \newline Conclusion of DRI: X \newline
% DS: (X $\lor$ Y) $\land$ $\lnot$X  $\vdash$ Y %(X ∨ Y) ∧ ¬X ⊢ Y
MP: (X $\to$ Y) $\land$ X $\vdash$ Y %(X → Y) ∧ X ⊢ Y
% MT: (X $\to$ Y) $\land$ $\lnot$Y  $\vdash$ $\lnot$X % (X → Y) ∧ ¬Y ⊢ ¬X
& \textbf{Context:} John and Mary are students in the same class. Usually students in their class do homework every day. John did not do his homework yesterday. Mary studied extra material last night. If Mary did her usual homework, she would have also reviewed her notes.\newline\newline \textbf{Question:} Can we conclude that Mary reviewed her notes last night? (Yes) \\ \midrule

\textbf{DRI\_MT} \newline\newline Conclusion of DRI: X \newline
% MP: (X $\to$ Y) $\land$ X $\vdash$ Y %(X → Y) ∧ X ⊢ Y
MT: (X $\to$ Y) $\land$ $\lnot$Y  $\vdash$ $\lnot$X % (X → Y) ∧ ¬Y ⊢ ¬X
& \textbf{Context:} Sara and David ordered dessert at a restaurant. Usually, people who order dessert also order coffee. Sara did not order coffee. David requested extra whipped cream. If a customer asks for extra toppings, it means they did not order coffee. \newline\newline \textbf{Question:} Can we conclude David asked for extra toppings? (No)     \\ \midrule

\textbf{PBD\_MP} \newline\newline Conclusion of PBD: X \newline 
% DS: (X $\lor$ Y) $\land$ $\lnot$X  $\vdash$ Y %(X ∨ Y) ∧ ¬X ⊢ Y
MP: (X $\to$ Y) $\land$ X $\vdash$ Y %(X → Y) ∧ X ⊢ Y
% MT: (X $\to$ Y) $\land$ $\lnot$Y  $\vdash$ $\lnot$X % (X → Y) ∧ ¬Y ⊢ ¬X
& \textbf{Context:} Jenny said the dog dug up the flower bed. Her brother said the dog did not dig up the flower bed. People usually tell the truth. Jenny is more trustworthy than her brother. If the dog dug up the flowers, it likely made a mess. \newline\newline \textbf{Question:} Can we conclude the dog made a mess? (Yes) \\ \midrule

\textbf{PBD\_MT} \newline\newline Conclusion of PBD: X \newline 
MT: (X $\to$ Y) $\land$ $\lnot$Y  $\vdash$ $\lnot$X %(X → Y) ∧ ¬Y ⊢ ¬X
& \textbf{Context:} John said the shirt was blue. Mary said the shirt was not blue. Normally people are correct when they make assertions. John had a closer look at the shirt than Mary. If the shirt was purple, it could not be blue. \newline\newline \textbf{Question:} Can we conclude the shirt was purple? (No) \\ \midrule

\textbf{REI\_MP} \newline\newline Conclusion of REI: X\newline
MP: (X $\to$ Y) $\land$ X $\vdash$ Y %(X → Y) ∧ X ⊢ Y
& \textbf{Context:} Ben, Mark, and Jacob took a history test. Students who study many hours usually pass history tests. Ben and Mark did not study many hours. If Jacob passed the history test, he must have paid attention in class. \newline\newline \textbf{Question:} Can we conclude Jacob paid attention in class? (Yes) \\ \midrule
            
\textbf{REI\_MT} \newline\newline Conclusion of REI: X\newline
% MP: (X $\to$ Y) $\land$ X $\vdash$ Y %(X → Y) ∧ X ⊢ Y
MT: (X $\to$ Y) $\land$ $\lnot$Y  $\vdash$ $\lnot$X %(X → Y) ∧ ¬Y ⊢ ¬X
& \textbf{Context:} John, Peter and Kate are students in math class. Students in math class normally do homework. John and Peter did not do their math homework. If Kate missed class then she did not do her math homework. \newline\newline \textbf{Question:} Can we conclude Kate missed class? (No) \\ \midrule

\textbf{REII\_MP} \newline\newline Conclusion of REII: X\newline
MP: (X $\to$ Y) $\land$ X $\vdash$ Y %(X → Y) ∧ X ⊢ Y
& \textbf{Context:} John bought a new phone. New phones usually come with a warranty. However, some new phones do not come with a warranty. If a phone has a warranty, then it has customer support.  \newline\newline \textbf{Question:} Can we conclude John's new phone has customer support? (Yes) \\ \midrule

\textbf{REII\_MT} \newline\newline Conclusion of REII: X\newline
% MP: (X $\to$ Y) $\land$ X $\vdash$ Y %(X → Y) ∧ X ⊢ Y
MT: (X $\to$ Y) $\land$ $\lnot$Y  $\vdash$ $\lnot$X %(X → Y) ∧ ¬Y ⊢ ¬X
& \textbf{Context:} Kate booked a room at hotel Y. Rooms at hotel Y are usually clean. There is at least one room at hotel Y that is not clean. If Kate's room has mold, then it is probably not clean. \newline\newline \textbf{Question:} Can we conclude Kate's room has mold? (No) \\

\bottomrule
\end{tabular}
}
\caption{Natural language examples of rule combinations of depth-2 for NM.}
\label{tab:NM_NL_Example_D2}
\end{table*}
% \usepackage[normalem]{ulem}
\begin{table*}
\centering
\resizebox{0.9\linewidth}{!}{
\begin{tabular}{m{10cm}|m{11cm}}
\toprule
Rule  & Examples \\ \midrule 

\textbf{Rule: d3\_1} \newline \newline 
\textbf{Assumptions:}\newline
1: A and B are objects of type T and have property S.\newline
2: Normally objects of type T with property S have property U.\newline
3: if A has property U implies C has property D\newline
4: if C has property D implies E has property F\newline \newline

\textbf{Question 1:} \newline Can we conclude if E does not have F then B has U?  (YES) \newline
\textbf{Question 2:} \newline Can we conclude if E does not have F then B does not have U?  (NO)

% \newline MP: (X $\to$ Y) $\land$ X $\vdash$ Y %(X → Y) ∧ X ⊢ Y
% Jim and Pam work at the same office. Normally, employees at that office get free lunch. Jim does not get free lunch. If Pam gets free lunch, then she gets an hour lunch break.",
%             "question": "Can we conclude Pam gets an hour lunch break?",

& \textbf{Context:} Smartphone A and Smartphone B both have GPS technology. Normally, smartphones with GPS technology also have internet connectivity. If smartphone A has internet connectivity, then Mike can access online maps. If Mike can access online maps, then Emily can get driving directions from Mike. \newline \newline \textbf{Question 1:} Can we conclude if Emily can not get driving directions from Mike, then smartphone B has internet connectivity? (Yes) \newline
\textbf{Question 2:} Can we conclude if Emily can not get driving directions from Mike, then smartphone B does not have internet connectivity? (No) 
\\ \midrule

\textbf{Rule: d3\_2} \newline \newline 
\textbf{Assumptions:}\newline
1: A and B are objects of type T and have property S. \newline    2: Normally objects of type T with property S have property U. \newline
3: if C has property G implies C has property D \newline
4: if A has property U implies E has property F \newline
5: either C has property G or E does not have property F or both \newline\newline

\textbf{Question 1:} \newline Can we conclude if C does not have D then B has U?  (YES) \newline
\textbf{Question 2:} \newline Can we conclude if C does not have D then B does not have U?  (NO)

% \newline MP: (X $\to$ Y) $\land$ X $\vdash$ Y %(X → Y) ∧ X ⊢ Y
% Jim and Pam work at the same office. Normally, employees at that office get free lunch. Jim does not get free lunch. If Pam gets free lunch, then she gets an hour lunch break.",
%             "question": "Can we conclude Pam gets an hour lunch break?",

& \textbf{Context:} Car A and car B are electric vehicles. Normally, electric vehicles (cars) have fast-charging capabilities. If car C is a hybrid, then car C has good fuel efficiency.	If car A has a fast-charging capability, then it implies that the environment is very eco-friendly. Either car C is a hybrid or the environment is not very eco-friendly, or both.\newline \newline 
\textbf{Question 1:} Can we conclude if car C is not a good fuel efficient then Car B has a fast-charging capability? (Yes) \newline
\textbf{Question 2:} Can we conclude if car C is not a good fuel efficient then Car B does not have a fast-charging capability? (No) 
\\ \midrule
\bottomrule
\end{tabular}
}
\caption{Natural language examples of rule combinations of depth-3 for NM.}
\label{tab:NM_NL_Example_D3}
\end{table*}
% \usepackage[normalem]{ulem}
\begin{table*}
\centering
\resizebox{0.9\linewidth}{!}{
\begin{tabular}{m{10cm}|m{11cm}}
\toprule
Rule  & Examples \\ \midrule 

\textbf{Rule: d4\_1} \newline \newline 
\textbf{Assumptions:}\newline
1: A and B are objects of type T and have property S. \newline
2: Normally objects of type T with property S have property U.\newline
3: if C has property G implies C has property D \newline
4: if E has property L implies E has property F \newline
5: either C has property G or E has property F or both \newline
6: if A has property U then E has property L \newline\newline
 
\textbf{Question 1:} \newline Can we conclude if C does not have D then B has U?  (YES) \newline
\textbf{Question 2:} \newline Can we conclude if C does not have D then B does not have U?  (NO)

% \newline MP: (X $\to$ Y) $\land$ X $\vdash$ Y %(X → Y) ∧ X ⊢ Y
% Jim and Pam work at the same office. Normally, employees at that office get free lunch. Jim does not get free lunch. If Pam gets free lunch, then she gets an hour lunch break.",
%             "question": "Can we conclude Pam gets an hour lunch break?",

& \textbf{Context:} Apple tree and Orange tree are fruit trees. Normally, fruit trees produce edible fruit. If Garden is regularly watered, then its plants are flourishing. If Orchard receives enough sunlight, then it yields high-quality fruit. Either Garden has regular watering or Orchard yields high-quality fruit or both. If the apple tree produces edible fruit, then Orchard receives enough sunlight. \newline \newline 
\textbf{Question 1:} Can we conclude if Garden does not have flourishing plants then the orange tree produces edible fruit? (Yes) \newline
\textbf{Question 2:} Can we conclude if Garden does not have flourishing plants then the orange tree does not produce edible fruit? (No) 
\\ \midrule

\textbf{Rule: d4\_2} \newline \newline 
\textbf{Assumptions:}\newline
1: A and B are objects of type T and have property S. \newline    2: Normally objects of type T with property S have property U. \newline
3: if C has property G implies C has property D \newline
4: if E has property L implies E has property F \newline
5: either C does not have property D or E does not have property F or both \newline
6: if A has property U then E has property L \newline\newline

\textbf{Question 1:} \newline Can we conclude if C has property G then B has U?  (YES) \newline
\textbf{Question 2:} \newline Can we conclude if C does not have G then B does not have U?  (NO)

% \newline MP: (X $\to$ Y) $\land$ X $\vdash$ Y %(X → Y) ∧ X ⊢ Y
% Jim and Pam work at the same office. Normally, employees at that office get free lunch. Jim does not get free lunch. If Pam gets free lunch, then she gets an hour lunch break.",
%             "question": "Can we conclude Pam gets an hour lunch break?",

& \textbf{Context:} Assume A and B are plants of species T and they both produce flowers. Normally, flowering plants of species T also bear fruit. If an animal C is a bird, then it can fly. If an environment has a lot of sunlight, then it supports plant growth. Either the bird cannot fly or the environment does not support plant growth or both. If plant A bears fruit, then the environment has a lot of sunlight. \newline \newline 
\textbf{Question 1:} \newline Can we conclude if the bird is capable of flying then plant B bears fruit? (Yes) \newline
\textbf{Question 2:} \newline Can we conclude if the bird can fly then plant B does not bear fruit? (No) 
\\ \midrule
\bottomrule
\end{tabular}
}
\caption{Natural language examples of rule combinations of depth-4 for NM.}
\label{tab:NM_NL_Example_D4}
\end{table*}
% \usepackage[normalem]{ulem}
\begin{table*}
\centering
\resizebox{0.9\linewidth}{!}{
\begin{tabular}{m{10cm}|m{11cm}}
\toprule
Rule  & Examples \\ \midrule \midrule

\textbf{Rule: d5\_1} \newline \newline 
\textbf{Assumptions:}\newline
1: A and B are objects of type T and have property S. \newline
2: Normally objects of type T with property S have property U. \newline
3: if C has property G implies C has property D \newline
4: if E has property L implies E has property F \newline
5: either C has property G or E has property F or both \newline
6: if I has property H then E has property L \newline
7: if A has property U then I has property H 
\newline\newline

\textbf{Question 1:} \newline Can we conclude if C does not have D then B has U?  (YES) \newline
\textbf{Question 2:} \newline Can we conclude if C does not have D then B does not have U?  (NO)

% \newline MP: (X $\to$ Y) $\land$ X $\vdash$ Y %(X → Y) ∧ X ⊢ Y
% Jim and Pam work at the same office. Normally, employees at that office get free lunch. Jim does not get free lunch. If Pam gets free lunch, then she gets an hour lunch break.",
%             "question": "Can we conclude Pam gets an hour lunch break?",

& \textbf{Context:} Rose and Lily are plants that flower. Normally, plants that flower also produce seeds. If a plant is a Cactus, and it has thorns, then it can survive in the desert. If a plant is an Orchid, and it has broad leaves, then it can grow in tropical areas. Either a Cactus has thorns, or an Orchid can grow in tropical areas, or both. If a Lotus has flowers, then an Orchid has broad leaves. If a Rose produces seeds, then a Lotus has flowers. \newline \newline 
\textbf{Question 1:} \newline Can we conclude if a Cactus cannot survive in the desert then a Lily produces seeds? (YES)\newline
\textbf{Question 2:} \newline Can we conclude if a Cactus cannot survive in the desert then a Lily does not produce seeds? (NO) 
\\ \midrule

\bottomrule
\end{tabular}
}
\caption{Natural language examples of rule combinations of depth-5 for NM.}
\label{tab:NM_NL_Example_D5}
\end{table*}

% Table \ref{tab:nm_examples} displays instances of general rules discussed in the paper by Lifschitz \cite{lifschitz1989benchmark}, specifically chosen for depth-1 non-monotonic logic. Out of the 11 default non-classical reasoning rules mentioned in the paper, we opted for 8. These include Default Reasoning with Several Defaults (DRS), Default Reasoning with Irrelevant Information (DRI), Default Reasoning with a Disabled Default ({DRD), Default Reasoning in an Open Domain (DRO), Reasoning about Unknown Expectations I (RE1), Reasoning about Unknown Expectations II (RE2), Reasoning about Unknown Expectations III (RE3), and Reasoning about Priorities (RAP). These rules constitute our selection for depth-1 non-monotonic logical reasoning. Moving on to depth-2, we integrated classical and non-classical logic. Table \ref{tab:NM_NL_Example_D2} outlines the combinations of rules prepared for the depth-2 logical reasoning task. In this context, we combined BDR, DRD, DRI, PBD, DRO, REII, and REIII from non-monotonic logic with MP, MT, and DS from propositional logic to form combinations for depth-2. Table \ref{tab:prompt_example_NM} shows a prompt that we have used to generate data instances for depth-2. The table shows an example of the BDR and DRD non-monotonic logic combined with the propositional logic - DS to generate depth-2 data. The instruction-based data generation can be seen in Table \ref{tab:prompt_example_NM}. 

Table \ref{tab:nm_examples} displays instances of general rules discussed in the paper by Lifschitz \cite{lifschitz1989benchmark}, specifically chosen for depth-1 non-monotonic logic.
Out of the 11 default non-classical reasoning rules mentioned in the paper, we opted for 8. These include Default Reasoning with Several Defaults (DRS), Default Reasoning with Irrelevant Information (DRI), Default Reasoning with a Disabled Default (DRD), Default Reasoning in an Open Domain (DRO), Reasoning about Unknown Expectations I (RE1), Reasoning about Unknown Expectations II (RE2), Reasoning about Unknown Expectations III (RE3), and Reasoning about Priorities (RAP).
These rules constitute our selection for depth-1 non-monotonic logical reasoning. Moving on to depths 2 through 5, we integrated classical and non-classical logic. Tables \ref{tab:NM_NL_Example_D2}, \ref{tab:NM_NL_Example_D3}, \ref{tab:NM_NL_Example_D4}, and \ref{tab:NM_NL_Example_D5} outline the combinations of rules prepared respectively for depth-2, depth-3, depth-4, and depth-5 logical reasoning tasks.
In this context, we combined BDR, DRD, PBD, DRO, REII, and REIII from non-monotonic logic with MP, MT, and DS from propositional logic to form combinations for depths 2 to 5 of data. Tables \ref{tab:prompt_example_NM}, \ref{tab:prompt_example_NM_3}, \ref{tab:prompt_example_NM_4}, and \ref{tab:prompt_example_NM_5} show the prompts that we used to generate data instances respectively for depths 2, 3, 4, and 5.
The instruction-based data generation can be seen in Tables \ref{tab:prompt_example_NM}, \ref{tab:prompt_example_NM_3}, \ref{tab:prompt_example_NM_4}, and \ref{tab:prompt_example_NM_5}. In addition to instruction-based generation, one-shot prompts were used for depth-3, depth-4, and depth-5 data generation as seen in Tables \ref{tab:prompt_example_NM_3}, \ref{tab:prompt_example_NM_4}, and \ref{tab:prompt_example_NM_5}.

\begin{table*}
\centering
\resizebox{0.9\linewidth}{!}{
\begin{tabular}{l}
\toprule  \midrule
\begin{tabular}[c]{@{}l@{}}

\textbf{Rule:} \\
\textbf{Assumptions:} \\ 
    1: A and B are objects of type T and have property P. \\
    2: Normally objects of type T with property P have property Q.\\
    3: A does not have property Q.\\
    4: If B has property Q then it implies B has property C.\\
\textbf{Question:} Can we conclude B has property C? \\\\

\textbf{Task 1:} \\Generate a short generic story that should only contain the natural \\language sentences for assumptions 1, 2, 3, and 4 using propositions to\\ replace the labels A, B and so on. \\The story should not include labels like p or q and so on. \\ \\

\textbf{Task 2:} \\Generate the question by replacing them with the entities with respective propositions. \\
\end{tabular}
\\ \midrule
\bottomrule
\end{tabular}
}
\caption{An example of prompt used to generate data instance for depth-2 using NM-BDR and PL-MP}
\label{tab:prompt_example_NM}
\end{table*}
\begin{table*}
\centering
\resizebox{0.9\linewidth}{!}{
\begin{tabular}{l}
\toprule  \midrule
\begin{tabular}[c]{@{}l@{}}

\textbf{Rule: d3\_1} \\ \\ 
\textbf{Assumptions:}\\
1: A and B are objects of type T and have property S.\\
2: Normally objects of type T with property S have property U.\\
3: if A has property U implies C has property D\\
4: if C has property D implies E has property F\\ \\

\textbf{Question 1:} \\ Can we conclude if E does not have F then B has U?  (YES) \\
\\
\textbf{Question 2:} \\ Can we conclude if E does not have F then B does not have U?  (NO) \\
\\
\textbf{Task 1:} Generate a short context paragraph by replacing all the entity labels A, B, \\and so on in the above context with propositions and real entities. The generated context\\ should have natural language sentences for all the sentences 1-4. It should not\\ include label representations like A or B and should not mention the words "property". \\
\\
\textbf{Task 2:} Generate questions 1 and 2 by replacing the respective labels from the generated context. \\\\
\textbf{Example 1:}\\
\textbf{Assumptions:}\\
Smartphone A and Smartphone B both have GPS technology.\\
Normally, smartphones with GPS technology also have internet connectivity.\\
If smartphone A has internet connectivity, then Mike can access online maps.\\
If Mike can access online maps, then Emily can get driving directions from Mike.\\\\
\textbf{Question 1:} \\Can we conclude if Emily can not get driving directions from Mike,\\ then smartphone B has internet connectivity? \\\\
\textbf{Question 2:}\\ Can we conclude if Emily can not get driving directions from Mike,\\ then smartphone B does not have internet connectivity?
\end{tabular}
\\ \midrule
\bottomrule
\end{tabular}
}
\caption{An example of prompt used to generate data instance for depth-3 for NM}
\label{tab:prompt_example_NM_3}
\end{table*}
\begin{table*}
\centering
\resizebox{0.9\linewidth}{!}{
\begin{tabular}{l}
\toprule  \midrule
\begin{tabular}[c]{@{}l@{}}

\textbf{Rule: d4\_1} \\ \\ 
\textbf{Assumptions:}\\
1: A and B are objects of type T and have property S. \\
2: Normally objects of type T with property S have property U.\\
3: if C has property G implies C has property D \\
4: if E has property L implies E has property F \\
5: either C has property G or E has property F or both \\
6: if A has property U then E has property L \\\\
 
\textbf{Question 1:} \\ Can we conclude if C does not have D then B has U?  (YES) \\
\\
\textbf{Question 2:} \\ Can we conclude if C does not have D then B does not have U?  (NO)
\\

\\
\textbf{Task 1:} Generate a short context paragraph by replacing all the entity labels A, B, \\and so on in the above context with propositions and real entities. The generated context\\ should have natural language sentences for all the sentences 1-4. It should not\\ include label representations like A or B and should not mention the words "property". \\
\\
\textbf{Task 2:} Generate questions 1 and 2 by replacing the respective labels from the generated context. \\\\
\textbf{Example 1:}\\
\textbf{Assumptions:}\\
Apple tree and Orange tree are fruit trees.\\
Normally, fruit trees produce edible fruit.\\
If Garden is regularly watered, then its plants are flourishing.\\
If Orchard receives enough sunlight, then it yields high-quality fruit.\\
Either Garden has regular watering or Orchard yields high-quality fruit or both.\\
If the apple tree produces edible fruit, then Orchard receives enough sunlight.\\
\\
\textbf{Question 1:} Can we conclude if Garden does not have flourishing \\plants then the orange tree produces edible fruit?\\
\\
\textbf{Question 2:} Can we conclude if Garden does not have flourishing\\ plants then the orange tree does not produce edible fruit?

\end{tabular}
\\ \midrule
\bottomrule
\end{tabular}
}
\caption{An example of prompt used to generate data instance for depth-4 for NM}
\label{tab:prompt_example_NM_4}
\end{table*}
\begin{table*}
\centering
\resizebox{0.9\linewidth}{!}{
\begin{tabular}{l}
\toprule  \midrule
\begin{tabular}[c]{@{}l@{}}

\textbf{Rule: d5\_1} \\ \\ 
\textbf{Assumptions:}\\
1: A and B are objects of type T and have property S. \\
2: Normally objects of type T with property S have property U. \\
3: if C has property G implies C has property D \\
4: if E has property L implies E has property F \\
5: either C has property G or E has property F or both \\
6: if I has property H then E has property L \\
7: if A has property U then I has property H 
\\\\

\textbf{Question 1:} \\ Can we conclude if C does not have D then B has U?  (YES) \\\\
\textbf{Question 2:} \\ Can we conclude if C does not have D then B does not have U?  (NO)
\\
\\
\textbf{Task 1:} Generate a short context paragraph by replacing all the entity labels A, B, \\and so on in the above context with propositions and real entities. The generated context\\ should have natural language sentences for all the sentences 1-4. It should not\\ include label representations like A or B and should not mention the words "property". \\
\\
\textbf{Task 2:} Generate questions 1 and 2 by replacing the respective labels from the generated context. \\\\
\textbf{Example 1:}\\
\textbf{Assumptions:}\\
Rose and Lily are plants that flower.\\
Normally, plants that flower also produce seeds.\\
If a plant is a Cactus, and it has thorns, then it can survive in the desert.\\
If a plant is an Orchid, and it has broad leaves, then it can grow in tropical areas.\\
Either a Cactus has thorns, or an Orchid can grow in tropical areas, or both.\\
If a Lotus has flowers, then an Orchid has broad leaves.\\
If a Rose produces seeds, then a Lotus has flowers.\\
\\
\textbf{Question 1:} Can we conclude if a Cactus cannot survive \\in the desert then a Lily produces seeds? (YES)\\\\
\textbf{Question 2:} Can we conclude if a Cactus cannot survive \\in the desert then a Lily does not produce seeds? (NO)

\end{tabular}
\\ \midrule
\bottomrule
\end{tabular}
}
\caption{An example of prompt used to generate data instance for depth-5 for NM}
\label{tab:prompt_example_NM_5}
\end{table*}

% Table \ref{tab:nm_examples} shows examples of generalized rules mentioned in the paper \cite{lifschitz1989benchmark}, we have selected for depth-1 non-monotonic logic. In total, we have selected eight rules out of 11 default non-classical reasoning rules mentioned in the paper. Default Reasoning with Several Defaults (DRS), Default Reasoning with Irrelevant Information (DRI), Default Reasoning with a Disabled Default (DRD), Default Reasoning in an Open Domain (DRO), Reasoning about Unknown Expectations I (RE1), Reasoning about Unknown Expectations II (RE2), Reasoning about Unknown Expectations III (RE3), Reasoning about Priorities (RAP) are the rules we have selected for depth one non-monotonic logical reasoning task. For depth-2, we have cross-combined the classical and non-classical logic. Table \ref{tab:NM_NL_Example_D2} shows all rule combinations prepared for the depth-2 logical reasoning task. We have combined BDR, DRD, DRI, PBD, DRO, REII, and REIII from NM logic and MP, MT, and DS from propositional logic to make combinations for depth-2. 
\section{Validation of Data Instances} \label{app:Data_Validation}
We involved four evaluators (who are also authors of this paper) for data validation. Each evaluator holds a graduate degree in computer science and has knowledge of logical reasoning. As discussed in Section \ref{sec:qualitative_analysis}, each sample is evaluated by one evaluator to ensure its logical correctness. We categorized errors into three distinct groups. The categories of errors identified are (i) Incorrect Logical Premises (ILP) which indicates that premises generated by the model in the context are logically incorrect (i.e., did not align with the intended conclusion), (ii) Leaking Conclusion (LC) where the context inadvertently revealed the conclusion, bypassing the need for the logical deduction, and (iii) Repetition of Samples (RS) where identical or nearly identical contexts are present, reducing dataset diversity. We found $\sim14.3\%$ (223 samples) of the total 1552 samples with ILP, $\sim3.7\%$ (57 samples) with LC, and $\sim3.7\%$ (57 samples) with RS. We mitigated all these errors manually from the generated data instances to provide a high-quality evaluation set. Furthermore, we also analyzed the number of samples we corrected for PL ($\sim~22\%$ - 115/525), FOL ($\sim19\%$ - 102/535), and NM ($\sim25.9\%$ - 127/492), highlighting the difficulty of generating instances for specific logics. Similarly, we also analyzed depth-wise instance correction where we corrected $\sim17.8\%$ (73/410), $\sim21\%$ (93/442), $\sim23.5\%$ (73/310), $\sim33.2\%$ (93/280), and $\sim21\%$ (23/110) for the depth $d_1$, $d_2$, $d_3$, $d_4$, and $d_5$, respectively, indicating the challenges of generating and validating multi-step reasoning context with increasing depth.

\section{Few shot evaluation Multi-LogiEval}
\label{app:fewshot_val}

We evaluate models in a few-shot setting (specifically, 3-shot) on Multi-LogiEval, revealing a notable enhancement in performance, as depicted in Table \ref{tab:few_shot_results}. In the 3-shot evaluation results, we observe notable improvements in the performance of various LLMs. GPT-4 consistently exhibits high accuracy across all depths, particularly excelling in PL and FOL. Though showing significant enhancements compared to its zero-shot performance across all the models, they still underperform in NM, highlighting a persistent challenge in this area. Open-source models such as Yi-34B and Mistral-7B, while benefiting from the 3-shot setup, still display noticeable performance drops in higher depths. Comparing these findings to the zero-shot results from Table \ref{tab:main_results}, we see a general trend of improved performance in the 3-shot setting, indicating the effectiveness of few-shot prompting. However, the observed performance drop from $d_4$ to $d_5$ in open-source models comparable across both settings, suggesting that while few-shot examples enhance overall accuracy, they do not fully mitigate the inherent challenges these models face in higher depths. Moreover, the performance trends identified in the zero-shot evaluation, such as the consistent decrease in accuracy for larger open-source models and the superior performance of proprietary models such as GPT-4 and ChatGPT in PL and FOL, remain similar in the 3-shot setting. 

%These results indicate the limitations and potential of current LLMs in logical reasoning, emphasizing the need for continued refinement and development to address the persistent gaps, particularly in handling non-monotonic reasoning tasks effectively.

% \usepackage{multirow}
\begin{table*}
\centering
\resizebox{\linewidth}{!}{
\begin{tabular}{c|c|c|c|c|c|c|c|c|c|c|c|c|c|c|c}
\toprule
\multirow{2}{*}{\textbf{Models}} & \multicolumn{5}{c|}{\textbf{Propositional}}                                                                                                                            & \multicolumn{5}{c|}{\textbf{First-Order}}                                                                                                                              & \multicolumn{5}{c}{\textbf{Non-Monotonic}}                                                                                                                                        \\ \cmidrule{2-16} 
                                 & \textbf{$d_1$} & \textbf{$d_2$} & \textbf{$d_3$} & \textbf{$d_4$} & \textbf{$d_5$} & \textbf{$d_1$} & \textbf{$d_2$} & \textbf{$d_3$} & \textbf{$d_4$} & \textbf{$d_5$} & \textbf{$d_1$} & \textbf{$d_2$} & \textbf{$d_3$} & \textbf{$d_4$} & \textbf{$d_5$} \\ \midrule
GPT-4                            & \multicolumn{1}{c|}{90.00}          & \multicolumn{1}{c|}{85.71}          & \multicolumn{1}{c|}{84.44}          & \multicolumn{1}{c|}{79.17}          & 73.33          & \multicolumn{1}{c|}{97.78}          & \multicolumn{1}{c|}{84.76}          & \multicolumn{1}{c|}{73.33}          & \multicolumn{1}{c|}{68.33}          & 73.33          & \multicolumn{1}{c|}{54.38}          & \multicolumn{1}{c|}{56.11}          & \multicolumn{1}{c|}{75.00}           & \multicolumn{1}{c|}{90.00}           & 75.00                                \\
ChatGPT                          & \multicolumn{1}{c|}{96.67}          & \multicolumn{1}{c|}{82.86}          & \multicolumn{1}{c|}{77.78}          & \multicolumn{1}{c|}{79.17}          & 80.00          & \multicolumn{1}{c|}{94.44}          & \multicolumn{1}{c|}{86.67}          & \multicolumn{1}{c|}{84.44}          & \multicolumn{1}{c|}{64.17}          & 64.44          & \multicolumn{1}{c|}{45.63}          & \multicolumn{1}{c|}{41.67}          & \multicolumn{1}{c|}{57.50}           & \multicolumn{1}{c|}{65.00}           & 45.00                                \\
Gemini                           & \multicolumn{1}{c|}{92.22}              & \multicolumn{1}{c|}{73.33}              & \multicolumn{1}{c|}{81.48}              & \multicolumn{1}{c|}{88.33}              & 77.78              & \multicolumn{1}{c|}{90.00}              & \multicolumn{1}{c|}{83.81}              & \multicolumn{1}{c|}{81.48}              & \multicolumn{1}{c|}{76.67}              & 57.78              & \multicolumn{1}{c|}{59.38}              & \multicolumn{1}{c|}{42.78}              & \multicolumn{1}{c|}{75.00}           & \multicolumn{1}{c|}{62.50}           & 75.00                                \\
Yi-34B                           & \multicolumn{1}{c|}{68.89}          & \multicolumn{1}{c|}{61.90}          & \multicolumn{1}{c|}{66.67}          & \multicolumn{1}{c|}{64.17}          & 64.44          & \multicolumn{1}{c|}{76.67}          & \multicolumn{1}{c|}{61.90}          & \multicolumn{1}{c|}{62.96}          & \multicolumn{1}{c|}{45.00}          & 51.11          & \multicolumn{1}{c|}{59.38}          & \multicolumn{1}{c|}{33.33}          & \multicolumn{1}{c|}{52.50}           & \multicolumn{1}{c|}{52.50}           & 50.00                                \\
Orca-13B                         & \multicolumn{1}{c|}{85.56}          & \multicolumn{1}{c|}{80.00}          & \multicolumn{1}{c|}{72.59}          & \multicolumn{1}{c|}{75.83}          & 68.89          & \multicolumn{1}{c|}{91.11}          & \multicolumn{1}{c|}{73.33}          & \multicolumn{1}{c|}{63.70}          & \multicolumn{1}{c|}{55.00}          & 42.22          & \multicolumn{1}{c|}{56.88}          & \multicolumn{1}{c|}{46.67}          & \multicolumn{1}{c|}{60.00}           & \multicolumn{1}{c|}{50.00}           & 50.00                                \\
Mistral-7B                       & \multicolumn{1}{c|}{80.00}          & \multicolumn{1}{c|}{64.76}          & \multicolumn{1}{c|}{71.11}          & \multicolumn{1}{c|}{73.33}          & 66.67          & \multicolumn{1}{c|}{93.33}          & \multicolumn{1}{c|}{71.43}          & \multicolumn{1}{c|}{62.96}          & \multicolumn{1}{c|}{62.50}          & 42.22          & \multicolumn{1}{c|}{37.50}          & \multicolumn{1}{c|}{36.11}          & \multicolumn{1}{c|}{45.00}           & \multicolumn{1}{c|}{57.50}           & 70.00                                \\ \midrule
\textbf{Avg}                     & \multicolumn{1}{c|}{\textbf{84.22}} & \multicolumn{1}{c|}{\textbf{75.05}} & \multicolumn{1}{c|}{\textbf{74.52}} & \multicolumn{1}{c|}{\textbf{74.33}} & \textbf{70.67} & \multicolumn{1}{c|}{\textbf{90.67}} & \multicolumn{1}{c|}{\textbf{75.62}} & \multicolumn{1}{c|}{\textbf{69.48}} & \multicolumn{1}{c|}{\textbf{59.00}} & \textbf{54.66} & \multicolumn{1}{c|}{\textbf{50.75}} & \multicolumn{1}{c|}{\textbf{42.78}} & \multicolumn{1}{c|}{\textbf{60.83}}  & \multicolumn{1}{c|}{\textbf{62.92}}  & \textbf{60.83}                       \\ \bottomrule
\end{tabular}
}
\caption{Few-shot Evaluation of LLMs in terms of accuracy on \textit{Multi-LogiEval}.}
\label{tab:few_shot_results}
\end{table*}
% \section{Extended discussion on Analysis} \label{app:Extended_Analysis}

\section{Extended first-order logic with n-ary relations}
\label{app:multi-variable fol}

First-order logic often involves handling n-ary relations involving more than two variables—such as the ternary relation in ``If $P(a,b,c) \land Q(c,d)$ then $R(a,d)$''. Moreover, one can alternate \textit{for all} ($\forall$), and \textit{there exists} ($\exists$) for any number of times in FOL, and that means there are an infinite number of such rules in first-order logic. As discussed in section \ref{sec 3:para:choice of inference rules}, our aim is not to build a comprehensive set covering all the possible inference rules but rather to evaluate the reasoning ability of language models up to a reasoning depth of five on a systematically curated set of inference rules. However, to evaluate the ability of LLMs to reason with such complex rules, we explore 7 such inference rules for which we generated data using a similar prompt structure as depicted in Figure \ref{fig:classical-logic-prompt-structure}. We generate 10 instances for each of the inference rule, resulting in 70 instances for evaluation. The choice of inference rules can be found in Table \ref{tab:multivariable_inference_rules}. We evaluate the large-scale models GPT-4, ChatGPT, and Gemini. These models achieve an average accuracy of $80\%$, $84.3\%$, and $90\%$, respectively. This demonstrates that these LLMs can comprehend multi-variable FOL, but the rules currently involve only single-step reasoning. Our work also shows that these models perform well with single-step reasoning. Exploring multi-step reasoning with multi-variable FOL presents an interesting direction for future research direction.
  
\section{Human Evaluation and Discussion}
\label{app:human_eval}

We have conducted a human evaluation on a subset of Multi-LogiEval. Specifically, we selected 15 unique instances covering all 5 depths (5 instances for each logic type) from Multi-LogiEval. This selection resulted in a total instances of 75 <context, question> pairs. We hired three graduate student volunteers to provide the evaluations. Task instructions provided to all three annotators are similar to prompts provided to LLMs. Each instance pair is answered/annotated by three different annotators with 0.853 inter-annotator agreement (measured with raw/observed agreement). Here are the results for three logic types averaged across three annotators for each depth. The average accuracies are $d_1$- 75.56, $d_2$ - 64.71, $d_3$ - 64.44, $d_4$ - 66.67, and $d_5$ - 64.44. From the results, we can observe that humans perform better at $d_1$ compared to higher depths. For higher depths, the human performance is low and consistent.

\paragraph{Discussion on Future Work} Our results provide deeper insights into LLMs' logical reasoning abilities by analyzing reasoning chains (Section \ref{subsec 4.3: Analysis and Discussion}) manually to some extent. Specifically, we analyzed where these models make mistakes and what are their limitations. We believe that such insights can help design better pre-training or alignment strategies to improve the reasoning abilities of LLMs. For instance, during pre-training, logical connectives can be treated differently at various stages of the transformer architecture. Additionally, in alignment techniques involving preference optimization, preference data can be created to prefer outputs with more logical correctness. The proposed approaches and findings in our paper can help in creating such datasets. Furthermore, exploring newer techniques such as DPO \cite{rafailov2024direct}, and KTO \cite{ethayarajh2024kto} for their suitability in improving logical reasoning also can be an interesting future direction.

\end{document}